\documentclass[acmsmall,screen]{acmart}

\usepackage{algorithm}
\usepackage{algorithmic}
\usepackage{graphicx}
\usepackage{comment}
\usepackage{comment}
\usepackage{booktabs}
\usepackage{subcaption}
\AtBeginDocument{%
  }

\setcopyright{none}
\renewcommand\footnotetextcopyrightpermission[1]{}
\acmDOI{}
\acmISBN{}
\acmConference[]{}{}{}
\acmBooktitle{}

\begin{document}

\title{An Enhanced Large Neighborhood Search Approach for the Capacitated Facility Location Problem with Incompatible Customers}
\author{Ida Gjergji}
\affiliation{%
  \institution{DBAI, TU Wien}
  \city{Vienna}
  \country{Austria}}
\email{ida.gjergji@tuwien.ac.at}

\author{Lucas Kletzander}
\affiliation{%
  \institution{DBAI, TU Wien}
  \city{Vienna}
  \country{Austria}}
\email{lucas.kletzander@tuwien.ac.at}

\author{Nysret Musliu}
\affiliation{%
  \institution{DBAI, TU Wien}
  \city{Vienna}
  \country{Austria}}
\email{nysret.musliu@tuwien.ac.at}

\author{Andrea Schaerf}
\affiliation{%
  \institution{University of Udine}
  \city{Udine}
  \country{Italy}}
\email{andrea.schaerf@uniud.it}



\begin{abstract}
A new variant of the classic capacitated facility location problem, which considers incompatibilities between customers, has recently been introduced in the literature. This problem captures the situation where given pairs of customers cannot be served by the same facility. Such a feature is crucial for many practical cases of location problems, such as the presence of hazardous or polluting materials and contention between competing costumers. In this paper, we propose a Large Neighborhood Search (LNS) method to solve this problem. Within the framework of LNS, we introduce three different destroy operators, which are combined in a hybrid manner, and we use an exact solver in the repair phase. Different algorithmic components are investigated for the design of LNS. The experimental analysis shows that our new method outperforms existing state-of-the-art metaheuristics, providing new best solutions for all available benchmark instances.
\end{abstract}

\keywords{Large Neighborhood Search, Facility Location Problem, Discrete Optimization}


\maketitle

\section{Introduction}

Facility location problems deal with determining optimal locations for facilities such as warehouses, factories, distribution centers, or service centers to meet the demands of customers or clients. These problems arise in various fields including logistics, supply chain management, telecommunications, and public services planning. The capacitated facility location problem (CFLP) extends the basic facility location problem \cite{LaNS15}, which is an NP-hard problem \cite{cornuejols1983uncapicitated}, by adding capacity constraints to facilities. The goal is to determine the optimal location of facilities while ensuring that the total quantity of goods served by each facility does not exceed its capacity.

In the \emph{multi-source} version of the capacitated problem (MS-CFLP) each customer can be served by a set of different facilities in order to collect its required quantity of goods. The alternative is the \emph{single-source} version, in which each customer is served entirely by a single facility. The \emph{multi-source capacitated facility location problem with customer incompatibilities} (MS-CFLP-CI) has been recently proposed by \citet{maia2023metaheuristic}, as an extension of MS-CFLP, in which there are pairs of customers that cannot be served by the same facility. This additional constraint may arise in situations where a customer does not want to share the supplying place with its competitors. Another application comes from materials subject to specific regulations or legal requirements that necessitate separate handling. Two challenging datasets have been proposed by \citet{maia2023metaheuristic} and by \citet{ceschia2024multi}, respectively. The instances of these datasets range from relatively small instances, with 50 facilities, to huge ones, up to 3000 facilities; and only the smaller ones, up to 150 facilities, have been solved to optimality.

Although advanced metaheuristic approaches \cite{maia2023metaheuristic,ceschia2024multi} were applied to these datasets, there is still room for improvement. For example, the optimal values, for instances up to 150 facilities, are not matched yet by the metaheuristic methods.


In this paper, which is a significant extension of the conference paper by \citet{gjergji2025large}, we introduce a novel solution approach for the MS-CFLP-CI based on Large Neighborhood Search (LNS). Our method
includes the design of new destroyers and a repair operator that relies
on an exact Mixed Integer Programming (MIP) solver. Additionally, several components of LNS are analyzed to enhance its performance. The proposed LNS, properly tuned in a statistically-principled manner, improves the state-of-the-art results for all instances of both available datasets. Summarizing, the main contributions of our paper are:
\begin{itemize}
    \item We investigate for the first time a Large Neighborhood Search approach for this problem.
    \item We propose novel destroy operators, such that their combination has been able to achieving state-of-the-art results.   
    \item We employ modern tuning methods to perform feature-based tuning and deeply analyze our approach including the impact of operators.  
    \item We devise and evaluate various configurations covering different algorithmic components of  the LNS framework.
    \item We tested our method on both publicly available datasets for the problem, which contain very large and challenging instances (up to 3000 facilities and 8000 customers). Our experiments show that our approach outperforms previous methods with statistical significance, by improving upper bounds for all existing instances in the literature.    
\end{itemize}

The paper is organized as follows: in Section \ref{sec:problem_definition} we give a formal definition of the MS-CFLP-CI problem. In Section \ref{sec:related_work} we discuss related work. Then, in Section \ref{sec:algorithm} we describe the components of our LNS, namely the initial solution, the  destroyers, and the repair operator. After analyzing the parameter configuration of LNS, in Section \ref{sec:experimental_ev} we present the experimental results, including the evaluation of different algorithmic components and the comparison with state-of-the-art methods. Finally, in Section \ref{sec:conclusions} we conclude our work and present ideas for future research. 


\section{Problem Definition}
\label{sec:problem_definition}
In the MS-CFLP-CI, there are $n$ customers to be served, each one with an associated demand $d_i$
$(i=1, \dots, n)$ to be satisfied. There are $m$ 
facilities that can be opened, each one with a capacity $s_j$ and an opening cost
$f_j$ ($ j = 1, \dots, m$). The cost of shipping one unit of good from facility $j$ to customer $i$ is denoted by $c_{ij}$. Finally, there is a set of pairs of incompatible customers $\mathcal{I}$, such
that for each $\langle i_1, i_2\rangle \in \mathcal{I}$, we state that $i_1$ and
$i_2$ cannot be served by the same facility.

The problem consists in deciding which facilities to open and the
quantities shipped to each customer from each (open) facility, in such a way that
the constraints on the capacity of the facilities are satisfied, the demands of the
customers are fully met, and the incompatibilities are respected. The objective function to be minimized is the sum of the opening costs and the shipping costs.

The mathematical model for the MS-CFLP-CI has been developed by \citet{ceschia2024multi}, reported here for self-containedness.

\begin{flalign}
 \min z =   \sum_{i=1}^n \sum_{j=1}^m c_{ij}x_{ij}d_i & + \sum_{j=1}^m f_j y_j \label{eq:of}	\\
   \sum_{j=1}^m x_{ij} = 1, &\quad i= 1, \dots, n \label{eq:supply} \\
    \sum_{i=1}^n d_ix_{ij} \leq s_j y_j, &\quad j=1, \dots, m \label{eq:capacity-open}\\
x_{ij} \leq w_{ij}, &\quad  i=1, \dots, n;~j= 1, \dots, m
\label{eq:channel}\\
w_{ij} \leq y_j, &\quad i=1, \dots, n;~j= 1, \dots, m \label{eq:redundant} \\
   w_{i_1j} + w_{i_2j} \leq 1, &\quad  \langle i_1, i_2\rangle \in \mathcal{I};~j = 1, \dots, m 
\label{eq:incompatibility}\\
y_j \in \{0,1\}, &\quad j=1, \dots, m \label{eq:y-domani}\\
 x_{ij} \in [0, 1], &\quad i=1, \dots, n;~j= 1, \dots, m \label{eq:x-domain}\\
 w_{ij} \in \{0, 1\}, &\quad i=1, \dots, n;~j= 1, \dots, m \label{eq:w-domani}
\end{flalign}

The decision variable $y_j$ takes value 1
if facility $j$ is open, and 0 otherwise, while the
decision variable $x_{ij}$ represents the fraction of the demand of
customer $i$ supplied by facility $j$.
The binary decision variables $w_{i j}$ are introduced to manage the incompatibility constraints,  so
that $w_{i j}$ is equal to 1 if the customer $i$ is supplied by
facility $j$ (even partially), 0 otherwise. Constraints (\ref{eq:incompatibility}) express the incompatibilities, whereas Constraints~(\ref{eq:channel}) link variables $w_{ij}$ to variables $x_{ij}$. Notice that Constraints (\ref{eq:redundant}) are redundant, as the rule that only open facilities can supply goods is already enforced by Constraints (\ref{eq:capacity-open}). They are customarily added given that they normally speed up convergence, by providing better cuts. 

As shown by \citet{ceschia2024multi}, the presence of incompatibilities makes the problem more complex in terms of both the computational time to solve it and the quality of the solutions. In fact, due to incompatibilities, the subproblem of finding the optimal supply from the open facilities is already NP-hard \cite{GoSp09}, whereas it is solvable in polynomial time in the basic case.

\section{Related Work}
\label{sec:related_work}

The literature on facility location is vast. We refer to the book by \citet{LaNS15} for a general and comprehensive introduction to the topic.

Restricting to the multi-source capacitated version of the problem (MS-CFLP), the most known and challenging benchmark is dataset \texttt{C} introduced by \citet{GuSp12}, with instances of 2000 facilities. The best known results on this dataset have been obtained by \citet{ABMR21} using flow techniques with an average gap from the lower bound of less than 0.1\%, improving on the previous best ones of the Benders decomposition by \citet{FiLS16}, which were about $1\%$ above on the most difficult instances.

Moving to our specific version of the problem (MS-CFLP-CI), as already mentioned, it  has been proposed by \citet{maia2023metaheuristic}, who also provided a dataset of $30$ instances, with size ranging from $50$ to $3000$ facilities, called \texttt{wlp} and available online\footnote{\url{https://github.com/MESS-2020-1/Instances}}. They propose four alternative metaheuristic approaches and compare them using two different time limits. In detail, the time limits proposed are based on the number of facilities $m$, and they are set to $10\sqrt{m}$ and $m$ seconds, respectively. All their methods were run on the same PC in order to have a fair comparison on a common ground. The method that turned out to be the best one is MineReduce, an approach based on data mining and iterated local search \cite{maia2020minereduce}. The MS-CFLP-CI has been subsequently dealt with by \citet{ceschia2024multi} using a Simulated Annealing approach. The key features of their method are the two-source restriction (i.e., maximum two suppliers per customer) and a pre-processing procedure that reduces the search space by restricting, based on the shipping costs, to a small subset of possible suppliers for each customer. The Simulated Annealing of \citet{ceschia2024multi} improved upon MineReduce of \citet{maia2023metaheuristic} on all instances, except for the smallest ones, with an average improvement of about $6\%$.
Using the model shown in Section~\ref{sec:problem_definition}, \citet{ceschia2024multi} computed the lower bounds for dataset \texttt{wlp}. However, they have been able to obtain them only for instances up to $400$ facilities, and strict ones (below $1\%$ gap) only up to $200$ facilities.
Finally, Ceschia and Schaerf also proposed a new dataset, called \texttt{cflp-ci} and available online\footnote{\url{https://github.com/iolab-uniud/ms-cflp-ci}}, composed of $50$ instances whose sizes range between $69$ and $2946$ facilities; and they provide their results on this dataset for both timeouts.


\section{Large Neighborhood Search for MS-CFLP-CI} 
\label{sec:algorithm}

Local search is an algorithmic paradigm based on the simple idea of traversing the
search space by iteratively stepping from one state to one of its neighbors.
Large Neighborhood Search (LNS) is a local search metaheuristic based on the idea of defining the neighborhood by means of one or more \emph{destroy} and \emph{repair} operators. 

LNS has been originally proposed by \citet{shaw1998using}, although similar approaches have also been considered independently (see the survey by \citet{ahuja2002survey} for alternative proposals). A comprehensive introduction to LNS has been provided by \citet{pisinger2019large}.

\begin{algorithm}[h]
\caption{LNS framework}
\label{lns_framework}
\begin{flushleft}
\textbf{Input}: Solution $s$, destroy operators $d_i(\cdot)\in\mathcal{D}$, repair operators $r_j(\cdot)\in\mathcal{R}$, acceptance criterion $a(\cdot,\cdot)$ \\
\textbf{Output}: Improved solution $s^*$ \\
\end{flushleft}
\begin{algorithmic}[1]
\STATE $s^* \gets s$
\WHILE{a stopping criterion is not reached}
\STATE Select $d_i$ from $\mathcal{D}$ and $r_j$ from $\mathcal{R}$
\STATE $s' \gets r_j(d_i(s))$
\IF {$a(s, s')$}
\STATE $s \gets s'$
\ENDIF
\IF {$s'$ better than $s^*$}
\STATE $s^* \gets s'$
\ENDIF
\ENDWHILE
\STATE \textbf{return} solution $s^*$
\end{algorithmic}
\end{algorithm}

The framework of LNS is given in Algorithm \ref{lns_framework}. 
In line $4$, a destroy operator shreds a part of the current solution and then a repair operator rebuilds it in a new way. In case of more than one operator of any kind, a strategy is needed for the selection of the operator in a given iteration in line $3$. The destroy operators are normally stochastic, whereas the repair operators may also be deterministic. The most common acceptance criterion, applied in line $5$, is accepting a solution if it is better than the current solution. When the acceptance criterion is fulfilled, $s'$ is accepted as the current working solution $s$ in line $6$, while the best overall solution is updated in line $9$.

\subsection{Solution Representation}
For Large Neighborhood Search, the solution is represented by a matrix that stores the quantity of goods shipped from each facility to each customer. We consider only solutions that satisfy the constraints: 
\begin{itemize}
    \item Goods are shipped only from open facilities.
    \item The demand of the customers is fully satisfied.
    \item The capacity of the facilities is never exceeded.
    \item Incompatible customers are not served by the same facility.
\end{itemize}

In order to speed up the search process, we make use of several auxiliary data structures. 
First, we have a set of open facilities $F$ and a set of closed facilities $G$. For any open facility $f \in F$, we have a list $C_{f}$ that contains all the customers assigned to this facility. On the other hand, for any customer $i$ we have a list $a_{i}$ with its assigned facilities and a second list $q_{i}$ indicating the corresponding quantity from these facilities.
\subsection{Initial Solution}
\label{sub:initial_solution}
We propose the following initial solution heuristic for the MS-CFLP-CI to provide a good starting solution to LNS. First, we sort the facilities in increasing order of their opening cost. Based on the cumulative capacity, we select enough facilities to fulfill the demand of the customers. As we have incompatibility constraints, a set of $k$ additional facilities is needed for the overall assignment. We checked the feasibility of the instances and initial cost for different values of $k$. For $k=0$ and $k=1$ infeasible solutions are reported, whereas for $k \in \{2, 3, 4, 5\}$ results are shown in Table \ref{init_solution_results}. From the obtained results, $k = 5$ provides the lowest total average cost for all instances without affecting the runtime. 
	
\begin{table*}[h]
\caption{Average total cost and time (in seconds) for different values of \texttt{k}}
\label{init_solution_results}
\centering
\begin{tabular}{ccccccccc}
\toprule
& \multicolumn{2}{c}{k=2} & \multicolumn{2}{c}{k=3} & \multicolumn{2}{c}{k=4} &  \multicolumn{2}{c}{k=5} \\
\cmidrule(lr){2-3} \cmidrule(lr){4-5} \cmidrule(lr){6-7} \cmidrule(lr){8-9}
& cost & time (s) & cost & time (s) & cost & time (s) & cost & time (s) \\
\midrule
avg  & $704094.99$ & $5.25$& $701488.79$ & $5.28$ & $699128.88$ & $5.27$&  \textbf{696518.04} & $5.26$ \\
\bottomrule
\end{tabular}
\end{table*}

Additionally, we investigate the order of assigning the customers to the open facilities. There are three strategies that we consider: uniformly, based on decreasing order of demand, and based on regret values. The notion of regret values is defined by  \citet{mulvey1984solving} as the absolute difference between the two closest open facilities for any customer \emph{i}. In our case, we compute the regret values as the absolute difference between the two cheapest open facilities in terms of shipping cost for all customers. The results show that the fastest performing strategy is the uniform one, while there are no significant differences in terms of total average cost among the three of them. For any customer \emph{i}, the open facilities are sorted in increasing order of shipping cost. Then the assigned quantity from the customer \emph{i} to the facility is the minimum between  customer's demand and facility's remaining capacity. We go through the sorted facilities until the demand of every customer has been met. This procedure is deterministic and we run it only one time per instance.

\subsection{Destroy Operators}
\label{sub:destroy_op}
A destroy operator defines a part of the solution \emph{s} which is subject to changes in an iteration, while the other portion of the solution remains intact. 
We refer to this part as the \emph{sub-problem}.
Initially, we considered operators from the literature defined for other versions of the facility location problem like random selection of facilities, swap operator (where facilities remain open but only re-assignment of customers, whenever possible, is allowed), high cost facilities (where facilities with higher cost are prioritized), and underloaded facilities (where facilities with higher unused capacities are prioritized). However, these operators did not yield good results. We believe this is because of the incompatibility constraints. This comprehensive investigation led to the definition of novel operators specifically designed for the MS-CFLP-CI.

In our case, the \textit{sub-problem} is defined in terms of the group of customers, the group of open facilities, and the group of closed facilities that are involved. 
In order to effectively define these elements of the sub-problem, we investigate the relationships between open facilities -- closed facilities, open facilities -- customers, and closed facilities -- customers. Note that the most important feature when we consider only customers is whether or not pairs of them are incompatible. In the repair phase, those customers can be reassigned to other facilities without affecting the feasibility of the solution as the incompatibilities are integrated in the constraints of the mathematical model used for the sub-problem.

The destroy operators proposed for the LNS are as follows, using a size parameter $\ell_{open}$:
\begin{itemize}
\item \texttt{Cheapest facilities (CF)}: For a randomly selected facility $f \in F$, identify its customers $C_{f}$. For any open facility $f'$, we define a measure of a quality $Q_{f'}$ as:
\begin{equation}
     Q_{f'} = \frac{\sum_{i \in C_{f}}c_{if'}}{\lvert C_{f} \rvert }  \quad \label{average unit cost}
\end{equation}
$Q_{f'}$ gives the average shipping cost of a facility for the set of customers $C_{f}$. The idea is to get the facilities that would in average provide service in a lower cost to the customers in $C_{f}$. The $\ell_{open}$ facilities with lowest $Q_{f'}$ and their customers are the elements of a sub-problem together with the initially selected facility $f$ and its customers $C_{f}$.
\item \texttt{Hybrid customers (HC)}: The design of this operator was inspired by the use of hybridization methods in the recombination stage in Evolutionary Algorithms \cite{blum2022hybridizations}. In our case, we recombine two destroy operators into one, by applying two different selections in the definition of a sub-problem. For a randomly selected facility $f$, identify its customers $C_{f}$. Other components of the sub-problem are chosen as follows:
\begin{itemize}
    \item \texttt{Cheapest customers (CC)}: Select $\left\lfloor \ell_{open}/2 \right\rfloor$ facilities whose customers have the lowest shipping cost with facility $f$. These facilities are denoted as $CC_f$. 
    \item \texttt{Expensive customer (EC)}: For the selected facility $f$ identify its most expensive customer in terms of shipping cost. Then, select $\left\lfloor \ell_{open}/2 \right\rfloor$ facilities that offer the lowest shipping cost for this customer, excluding $f$, other facilities offering service to this customer (if any), and $CC_f$ facilities. These facilities are denoted as $EC_f$.
\end{itemize}
In the sub-problem we then have facility $f$ (also including other facilities that serve the expensive customer) and its customers $C_{f}$ joined with the chosen facilities $CC_f$, $EC_f$, and their assigned customers.
\end{itemize}

Another component of the sub-problem is the group of closed facilities. Preliminary results show that it is favorable to also have closed facilities using the parameter $\ell_{closed}$ in each iteration so as to check whether they are better candidates for the group of selected customers compared to the involved open facilities. Additionally, as the quality measure in Equation \eqref{average unit cost} has shown beneficial impact, we also use it for defining one of the techniques for the closed facilities, to ensure good results with all destroy operators. There are two strategies that we consider for the collection of the closed facilities, which are chosen randomly according to a probability parameter $p_{MC}$:
\begin{itemize}
    \item \texttt{Random facilities(R)}: Select $\ell_{closed}$ random closed facilities from $G$.
    \item \texttt{Minimum cost facilities (MC)}: Get $\ell_{closed}$ closed facilities that offer minimum $Q_g$,  where $Q_g$ is calculated for any closed facility $g \in G$ as in Equation \eqref{average unit cost} considering all customers of the defined sub-problem.  
\end{itemize}

\subsection{Repair Operator}
\label{sub:repair_op}
In the repair phase, the destroyed part of the solution is restored. To achieve this, we declare the model as given in Section \ref{sec:problem_definition} for the sub-problem using the MIP solver Gurobi. For the components of the sub-problem all changes are allowed: open facilities can remain open or get closed, closed facilities can remain closed or be opened, and customers are allowed to change their previously assigned facilities. The sub-problem requirements are the same as for the master problem.

A crucial element of integrating an exact solver in the repair stage is the time it needs for an iteration. Declaring the model and solving it in an iterative fashion can be extremely time consuming. To avoid this, we carefully examine the possible options to provide a more compact model. Regarding this matter, we add one constraint that limits the number of facilities that can be open at the end of an iteration. If there are $o$ open facilities involved in the given iteration, then the extra constraint allows at most $o+2$ open facilities after the repair stage. From initial results, we could see that usually no more than $2$ additional facilities are opened. Constraining on the number of possible facilities to open is beneficial in terms of the runtime that the repair stage takes.

Furthermore, at each step we provide to Gurobi a \emph{cutoff} value, which is an upper limit for the solution cost. The cutoff value is the initial cost of the sub-problem that is considered in the iteration. This action helps to speed up the search process.

\section{Experimental Evaluation}
\label{sec:experimental_ev}
The experiments for tuning, ablation analysis, and testing algorithmic components of LNS were run on a computing cluster, equipped with two Intel Xeon E5-2650v4 @ ${2.20}$ CPUs with $12$ cores and on a single thread. Previous results by both \citet{maia2023metaheuristic} and  \citet{ceschia2024multi}, which represent state-of-the-art methods for MS-CFLP-CI,  have been obtained on an AMD Ryzen ThreadripperPRO 3975WX with 32 cores (3.50 GHz), with 64 GB of memory and running Ubuntu Linux 22.4, on a single core. For a direct comparison, we were able to run the LNS experiments on the same machine. The proposed LNS is compared with other methods based on the percentage gap value defined as follows:
\begin{equation}
    \mathrm{GAP_{sol}} = \frac{(Z_{sol}-\mathrm{BKS})}{\mathrm{BKS}}\cdot100 \label{gap equation}
\end{equation}

In Equation \eqref{gap equation}, $Z_{sol}$ is the objective function value of the solution, and BKS represents the best-known solution reported in the literature. The LNS approach is compared with all methods from \citet{maia2023metaheuristic} and \citet{ceschia2024multi}: MineReduce-based Multi-Start Iterated Local Search (MR-MS-ILS), Greedy Randomized Adaptive Search Procedure (GRASP), Permutation-coded Evolutionary Algorithm (PcEA), Multi-start Greedy (MG), and Simulated Annealing (SA). In the experimental evaluation we consider the dataset \texttt{wlp} by \citet{maia2023metaheuristic} and the dataset \texttt{cflp-ci} by \citet{ceschia2024multi}, with a  total of $80$ instances. Two different timeouts are used to assess the performance of the algorithms, $10\sqrt{m}$ seconds and $m$ seconds, where $m$ denotes the number of facilities in the corresponding instance. The BKS for $78$ out of $80$ instances are achieved by Simulated Annealing, and the other $2$ by MineReduce. 
As our LNS method, as well as the other methods, is non-deterministic, we report like the others the results of 10 runs and for both timeouts, using different seeds, which are recorded for reproducibility purposes.

\subsection{Exact Approach}
Before investigating LNS for the MS-CFLP-CI, we verified whether we could solve 
the problem by using Gurobi directly on the mathematical model given in Section \ref{sec:problem_definition} for all available instances. 
Similar analysis was done by \citet{ceschia2024multi} using CPLEX, but  only for dataset \texttt{wlp}. The time budget set for these experiments was $7200$ seconds. Our results are shown in Table \ref{tab:exact_solvers}, where we compare the results from CPLEX \cite{ceschia2024multi} and our results using Gurobi. The entries in bold indicate the solutions that both CPLEX and Gurobi report as optimal. For other instances, each entry is the objective function value that the solver delivers at the end of the run. Note that for the majority of the instances no results are reported as they are beyond the capacities of exact solvers. This is due to their size that causes the execution to run into a memory error. Regarding Table \ref{tab:exact_solvers}, the largest instance that Gurobi can solve to optimality has $150$ facilities, and it can only give feasible solutions for instances with no more than $500$ facilities with a runtime of $7200$ seconds. Moreover, for the instances that it can handle, Gurobi gives better bounds for $7$ instances compared to CPLEX. As for the MS-CFLP-CI problem instances range up to $3000$ facilities, the use of alternative methods such as metaheuristics is necessary to tackle such large instances.
\begin{table}[h]
\caption{Results from CPLEX and Gurobi}
\label{tab:exact_solvers}
\centering
\begin{tabular}{crrrr}
\toprule
instance & $m$ & $n$ & CPLEX & Gurobi\\
\midrule
wlp01 & 50 & 115 & \textbf{28716} & \textbf{28716} \\
wlp02 & 100 & 253 & \textbf{52952} & \textbf{52952} \\
wlp03 & 150 & 345 & \textbf{64296} & \textbf{64296} \\
wlp04 & 200 & 479 & $84633$ &  $84633$ \\
wlp05 & 250 & 601 & $107323 $ &  $103857$ \\
wlp06 & 300 & 705 & $115295$ &  $111654$ \\
wlp07 & 400 & 1012 & $170100$ &  $162277$ \\
wlp08 & 500 & 1277 & $-$ &  $187938$ \\
wlp21 & 75 & 172 & \textbf{38067} &  \textbf{38067} \\
wlp22 & 175 & 428 & $74473$ &  $74469$ \\
wlp23 & 275 & 694 & $124991$ &  $119091$ \\
wlp24 & 450 & 1128 & $176721 $ &  $168816$ \\
cflp-ci-11 &69  & 156 & &  \textbf{30728} \\
cflp-ci-39 & 383 & 1131 & & $189284$ \\
\bottomrule
\end{tabular}
\end{table}

\subsection{Parameter Configuration}
\label{sub:param_configuration}

The parameters considered for the LNS algorithm are the sub-problem size as $\nu$, the number of closed facilities as $\ell_{closed}$, the probability of selecting minimum cost closed facilities (\texttt{MC}) as $p_{MC}$ (the probability of random selection of closed facilities (\texttt{R}) is $p_{R}=1-p_{MC}$), the weight of the operator \texttt{CF} as $p_{CF}$ (where the weight of \texttt{HC} is $p_{HC} = 1 - p_{CF}$), and the time allocated to Gurobi in the repair stage as $t_{g}$. The time given to Gurobi $t_{g}$ has been determined by manual trials. Based on the design of the repair phase, we fix this value to $t_{g}=20$ seconds. For the other parameters, we use the automatic parameter configuration tool \texttt{irace} \cite{LOPEZIBANEZ201643}. The instances used for the tuning stage are \texttt{wlp01}-\texttt{wlp20}. This subset of $20$ instances was also used during the tuning procedure in the methods introduced by \citet{maia2023metaheuristic}. As the instance size varies considerably across instances, we perform feature-based tuning. Therefore, we define two separate tuning sets by conditioning on the respective number of facilities ($m$) of each instance. In the first tuning set there are instances with up to $700$ facilities (\texttt{wlp01}-\texttt{wlp10}) and in the second tuning set there are instances with more than $700$ facilities (\texttt{wlp11}-\texttt{wlp20}). For each case, we use a budget of $5000$ experiments in the tuning process with a timeout of $300$ seconds per experiment. Additionally, we provide to \texttt{irace} an initial configuration, which contains the same parameter values for both tuning sets. The parameter values, including their range and tuned values output by \texttt{irace}, are displayed in Table \ref{tab:parameter_setting}. For $\ell_{closed}$, $p_{MC}$, $p_{CF}$ there are no significant differences between the two tuning sets in the values recommended by \texttt{irace}. Conversely, the parameter $\nu$ for instances with more than $700$ facilities is almost half of the $\nu$ value for the smaller instances. The sub-problem size is an essential parameter of the algorithm. Indeed, a high value of $\nu$ can cause the repair stage to be really slow as it needs more time to declare and to solve a problem for more customers and facilities. On the other hand, a small value of $\nu$ can lead to fewer and smaller improvements. Thus, it is mandatory to provide a value of $\nu$ that works well for all of the instances. The destroy operator size $\ell_{open}$ is then calculated as defined by Equation~\eqref{eq:dl}, where we approximate the number of facilities that would offer service to $\nu$ customers by using the ratio of the current number of open facilities to the number of customers:
\begin{equation}
\label{eq:dl}
    \ell_{open}=\frac{ \nu \cdot {\lvert F \rvert }}{n}
\end{equation}

We use the parameters $\nu$, $\ell_{closed}$, $p_{MC}$, and $p_{CF}$ based on the instance size for our experiments as recommended from \texttt{irace}. The \emph{roulette wheel} approach is utilized to select the destroy operators throughout the LNS procedure.  

\begin{table}[t]
\caption{Parameter values used for the LNS}
\label{tab:parameter_setting}
\centering
\begin{tabular}{ccccc}
\toprule
Parameter & Description & Domain & \multicolumn{2}{c}{Tuned} \\
\cmidrule(lr){4-5}
 & & & $m \leq 700$ & $ m > 700$ \\
\midrule
$\nu$ & sub-problem size & $[10, 100]$ & $65$ & $35$ \\
$\ell_{closed}$ & number of closed facilities & $[5, 30]$ & $9$ & $6$ \\
$p_{MC}$ & probability to use \texttt{MC} & $[0, 0.5]$ & $0.44$ & $0.45$ \\
$p_{CF}$ & probability to use \texttt{CF} & $[0, 0.8]$ & $0.34$ & $0.35$ \\
\bottomrule
\end{tabular}
\end{table}

\subsection{Investigating Different Components of LNS}
For a deeper analysis of the proposed LNS, we investigate several of its components. Different configurations are formulated, each of which addresses a specific algorithmic element of LNS.
\subsubsection{Ablation Analysis}
\begin{figure}[h]
\centering
\begin{minipage}[b]{0.45\textwidth}
\centering
\includegraphics[width=\linewidth]{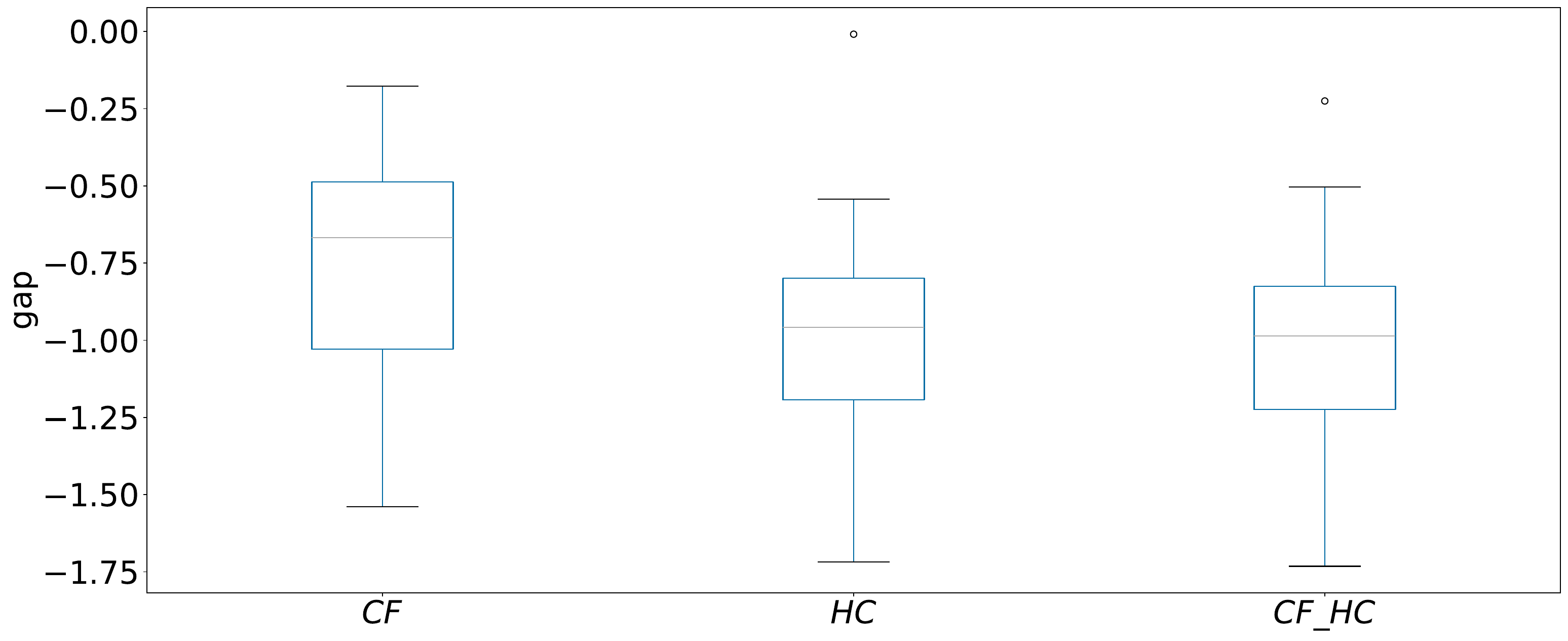}
\caption{GAP values for all the evaluated operator configurations}
\label{fig:ablation_gap}
\end{minipage} \hfill
\begin{minipage}[b]{0.45\textwidth}
\centering
\includegraphics[width=\linewidth]{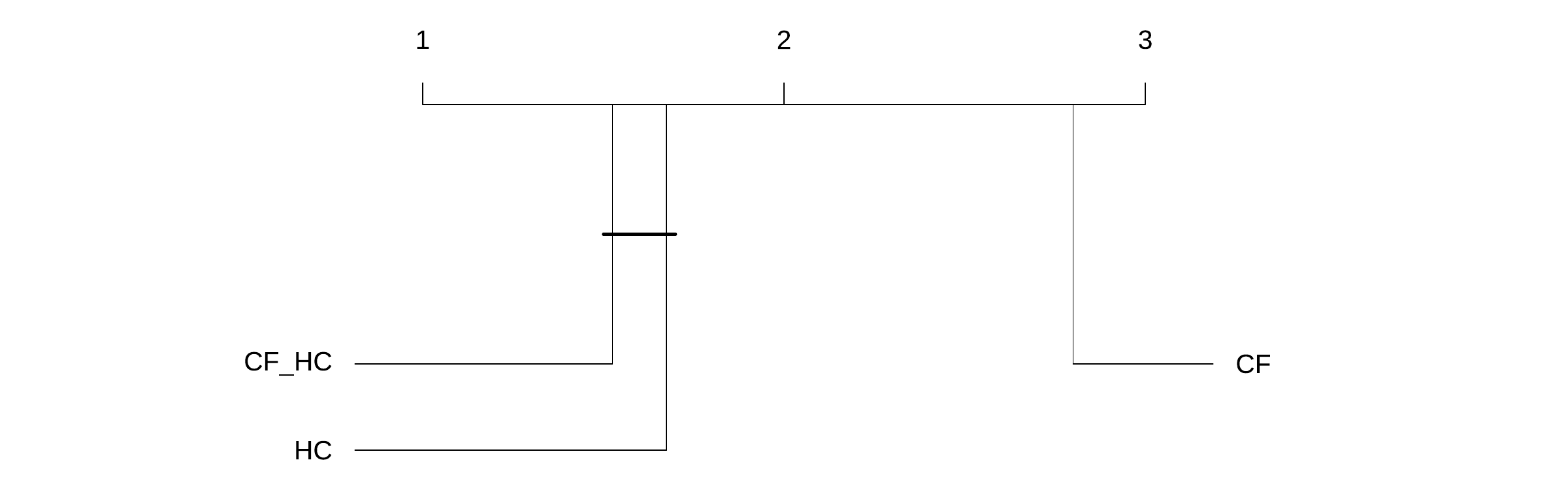}
\caption{Critical difference plot for all the evaluated operator configurations}
\label{fig:ablation_cd}
\end{minipage}
\end{figure}

We investigate the performance of the designed destroy operators by performing an ablation analysis. We consider each destroy operator individually \texttt{CF}, \texttt{HC}, and their combination \texttt{CF}\,$+$\,\texttt{HC}. These scenarios use all parameters as reported in Table \ref{tab:parameter_setting}, except $p_{CF}$ and $p_{HC}$. For these configurations the gap values from the average results of $10$ runs are shown in Figure \ref{fig:ablation_gap}. The first observation from the ablation analysis is all configurations provide negative gap values, which indicate that new solutions have been found. In order to better assess the performance of each configuration, we also perform statistical tests using the R script \texttt{scmamp} \cite{calvo2016scmamp}. Based on the average results for these $3$ configurations, the Friedman test gives a p-value smaller than $2.2 \times 10^{-16}$ which shows that not all of them have the same performance. Furthermore, we apply the Nemenyi post-hoc test which gives the ranking of these configurations based on their average performance.

According to these results, the configurations are ranked from the best to the worst performing configuration in the subsequent manner: \texttt{CF\,+\,HC}, \texttt{HC}, and \texttt{CF}. Figure~\ref{fig:ablation_cd} shows the results in a critical difference plot. Here, a lower index (given on top) shows better performance. Horizontal bars (e.g., between \texttt{CF\,+\,HC} and \texttt{HC}), indicate that no significant difference is given, while methods without a direct bar connection (e.g., \texttt{CF} and \texttt{HC}) show significant difference. As displayed in Figure \ref{fig:ablation_cd}, no statistically significant difference is obtained between the performance of \texttt{HC} and \texttt{CF\,+\,HC}. Such results are consistent with the reported values from the tuning process, attributing to the \texttt{HC} operator a higher weight compared to the \texttt{CF} operator. However, as the configuration \texttt{CF\,+\,HC} is first ranked and provides lower minimum values compared to \texttt{HC}, we use this configuration for further experiments.


 To evaluate other configurations, we perform $10$ runs in each case and the resulting average gap values are compared with the average gap values of the original LNS. 
These configurations are as follows: 
\subsubsection{Changing the initial solution:}
\begin{figure}[h]
\centering
\includegraphics[width=0.5\linewidth]{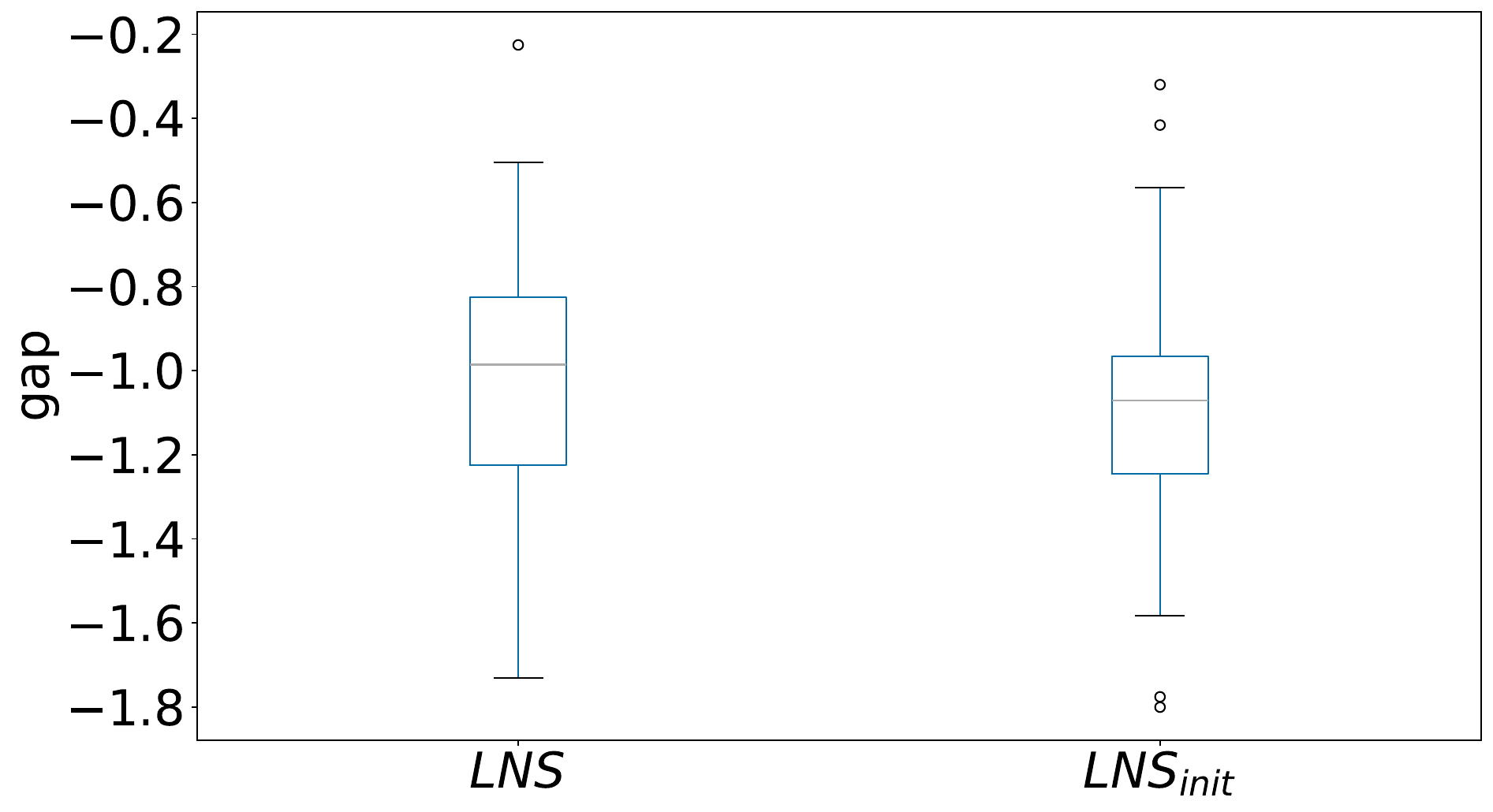}
\caption{GAP values for LNS and LNS$_{init}$}
\label{fig:init_baseline}  
\end{figure}

LNS$_{init}$. We provide as initial solution to LNS the greedy algorithm in SA \cite{ceschia2024multi}, which selects at each stage the $\langle$facility, customer$\rangle$ pair with the minimum cost, taking into account the opening cost in an amortized way, and assigning the maximum quantity that can be supplied from the facility to the customer. This configuration allows LNS to start the local search procedure from the same solution as SA. The results of LNS$_{init}$ and LNS are presented in Figure \ref{fig:init_baseline}, where LNS$_{init}$ achieves improvements compared to LNS. This is because starting from a better solution allows LNS$_{init}$ to make more effective use of the available time to search for further improvements.

\subsubsection{Changing the repair stage:}
\begin{figure}[h]
\centering
\includegraphics[width=0.5\linewidth]{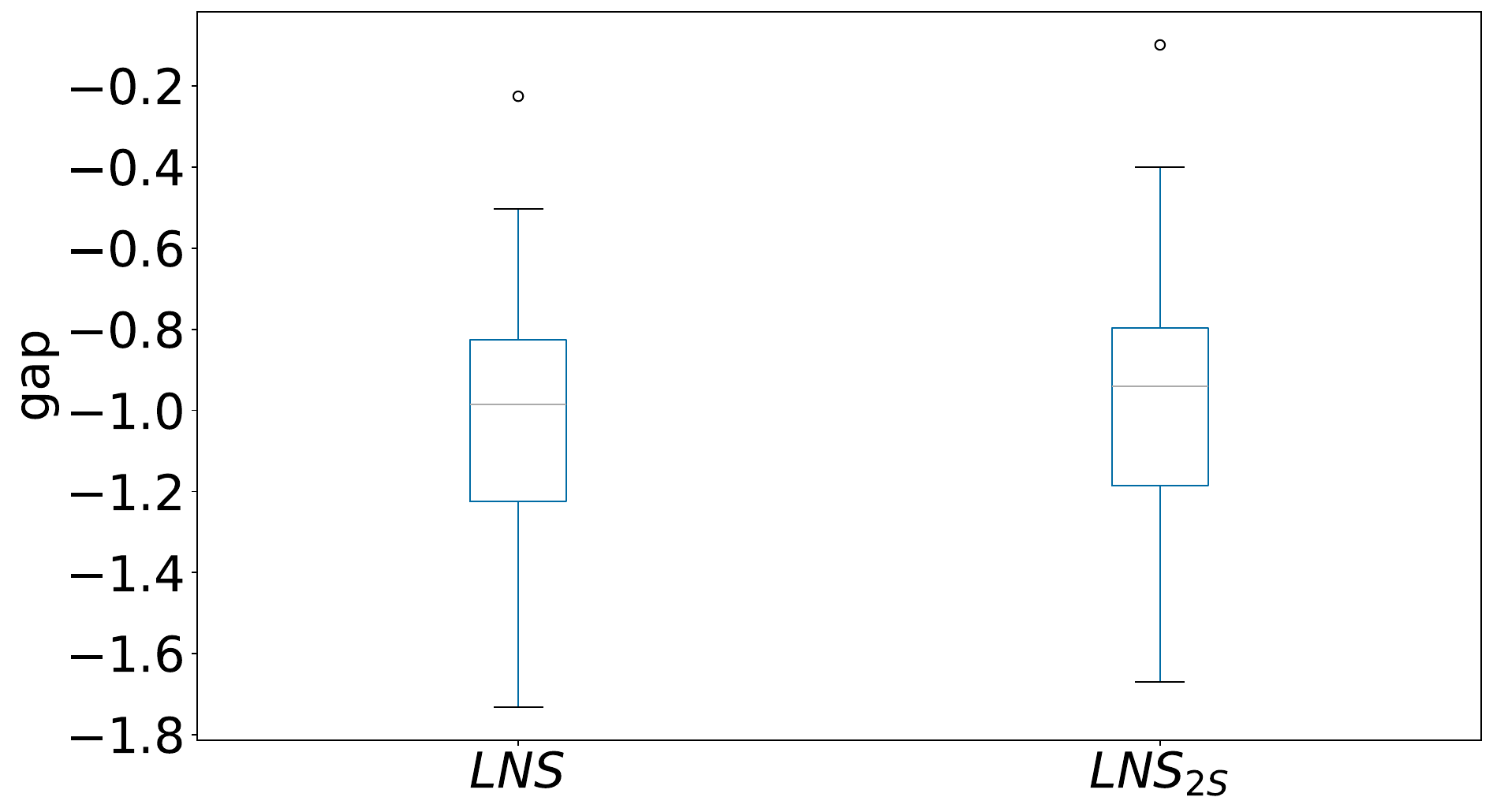}
\caption{GAP values for LNS and LNS$_{2S}$}
\label{fig:2s_baseline}  
\end{figure}

LNS$_{2S}$. Further constraints are declared in the repair stage such that each customer can be supplied by at most $2$ facilities. For a set of facilities $F'$ and a set of  customers $C'$ involved in the sub-problem, the following constraints are added to the model used in the repair phase:
    \begin{equation}
        \sum_{j \in F'} w_{ij} \leq 2, \quad i\in |C'| 
    \end{equation}
Note that these extra constraints do not imply that LNS offers a \texttt{2S} restriction, like SA, because in the sub-problem it is allowed to consider partial demand of customers. In Figure \ref{fig:2s_baseline}, the results of LNS$_{2S}$ are compared with LNS and no improvements are obtained with the change in the repair stage.
\subsubsection{Changing the acceptance criterion:} 
\begin{figure}[h]
\centering
\includegraphics[width=0.5\linewidth]{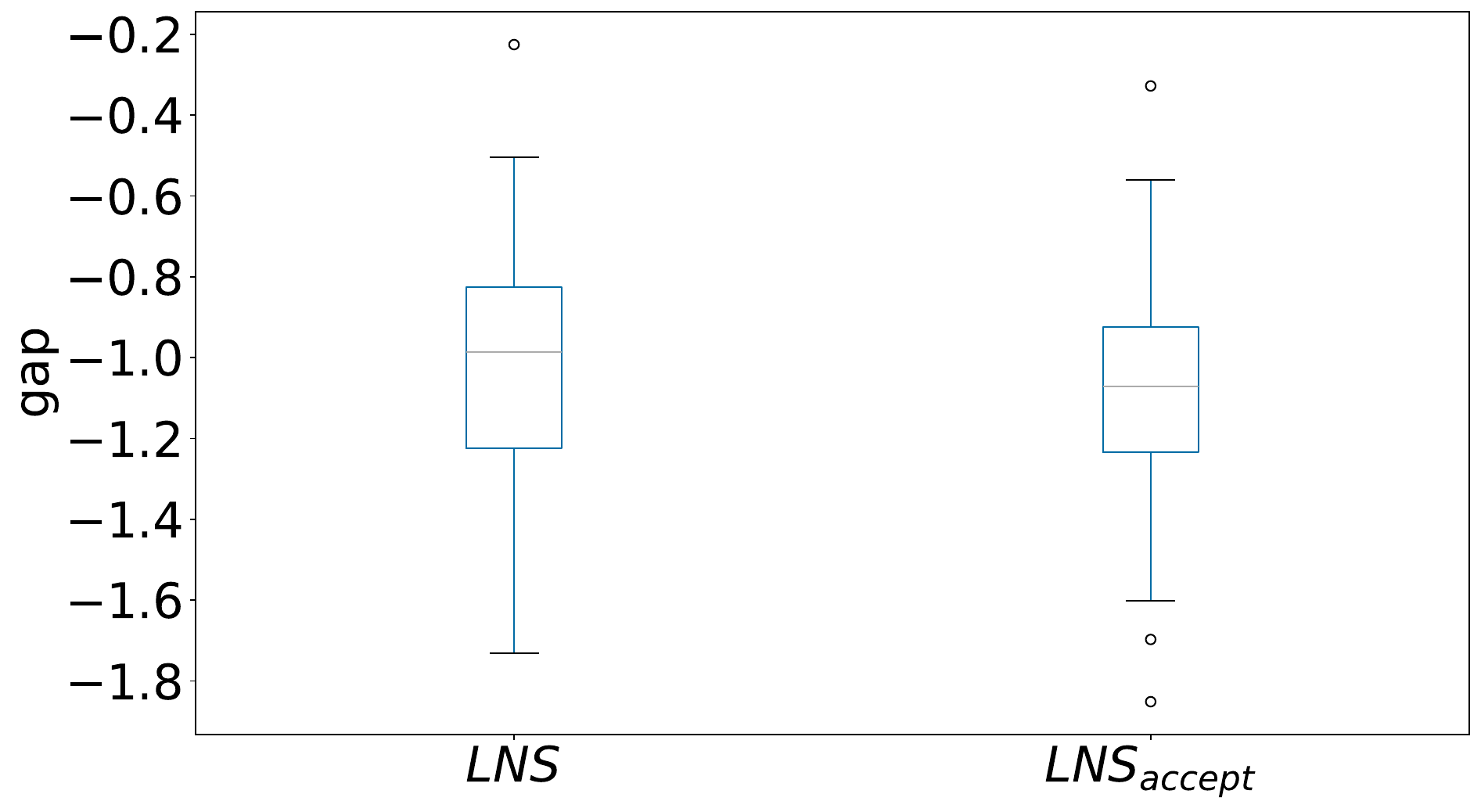}
\caption{GAP values for LNS and LNS$_{accept}$}
\label{fig:accept_baseline} 
\end{figure}

LNS$_{accept}$. In our LNS implementation, the sub-problem cost is used in the repair stage as a cutoff value. 
This implies that the repaired solution is always improving, otherwise the current one is retained. In the LNS$_{accept}$ configuration, we also accept solutions with the same cost value. 
Therefore, the structure of the solution is allowed to change in the repair stage even when no better solution is found.
This can help avoid getting stuck in local optima. The results of this configuration are shown in Figure \ref{fig:accept_baseline} and LNS$_{accept}$ provides better solutions in comparison to LNS.

\subsubsection{Dynamically changing the operators' weight:} 
\begin{figure}[h]
\centering
\includegraphics[width=0.5\linewidth]{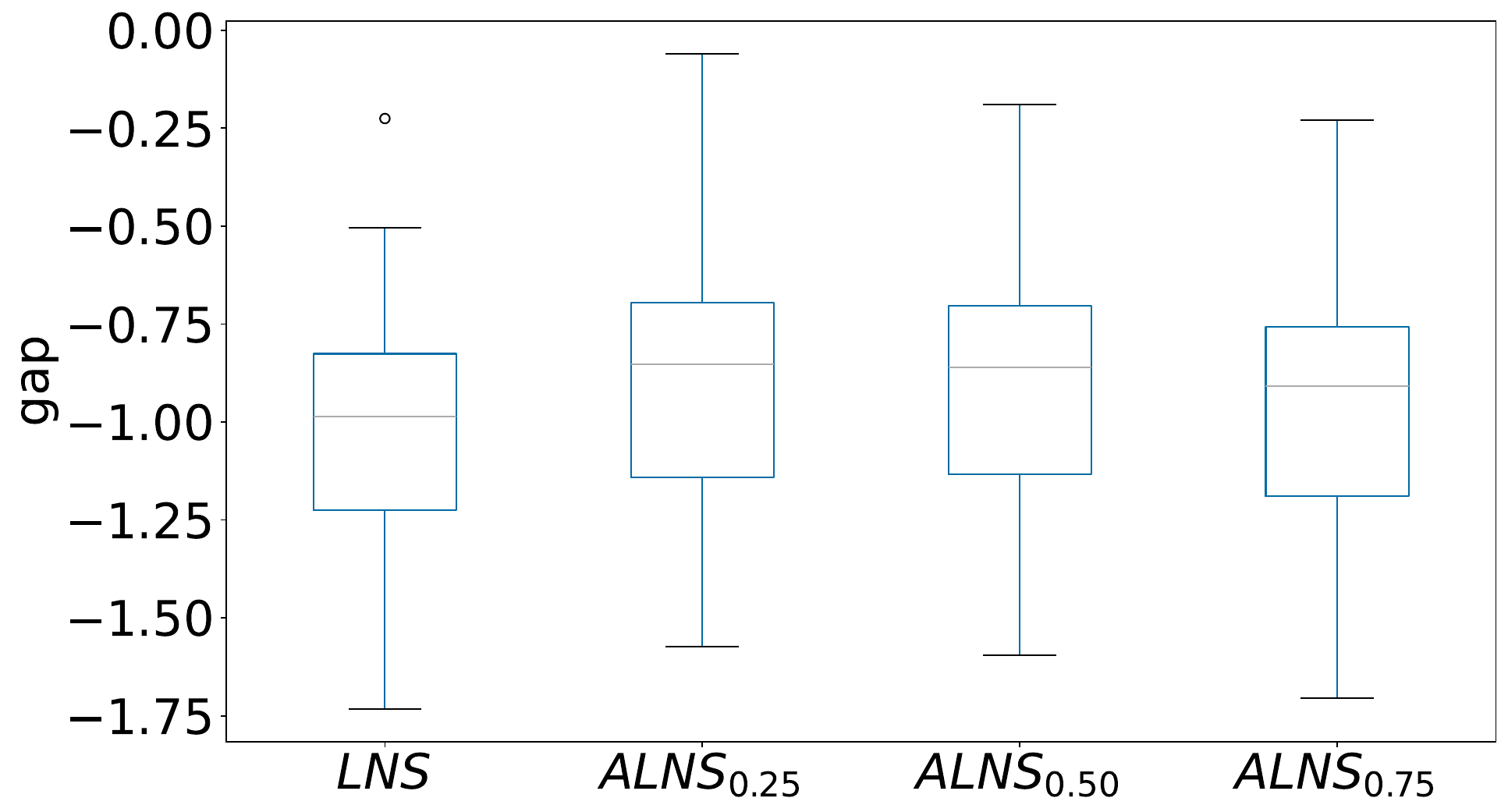}
\caption{GAP values of LNS and ALNS}
\label{fig:alns_baseline}    
\end{figure}

ALNS. Updating the weights of the destroy operators during the execution of the algorithm known as Adaptive Large Neighborhood Search \cite{ropke2006adaptive}. As outlined by  \citet{turkevs2021meta} it is important to investigate if adding the adaptive layer to the LNS can further enhance the performance, although this doesn't always improve the results. Initially, the weights of two destroy operators are equal $p_{d_i} = \frac{1}{|D|}$. Then in an iteration, each of proposed operators can be selected according to the roulette wheel approach:
    \begin{equation}
        P_{d_i} = \frac{p_{d_i}}{\sum_{v=1}^{|D|}p_{d_v}}
    \end{equation}
    The weights of the operators are updated in each iteration using the weight update function by \citet{mazzoli2024investigating}:
    \begin{equation}
        p^{it+1}_{d_i} = \lambda p^{it}_{d_i} + (1-\lambda) \frac{\sum_{j=0}^{it} selection^j_{d_i}}{\sum_{j=0}^{it}t^j_{d_i}}
    \end{equation}
$selection^j_{d_i}$ is 1 if applying operator $d_i$ in iteration $j$ results in an improved solution and $0$ otherwise. $t^j_{d_i}$ is the time that operator $d_i$ took in iteration $j$ ($t^j_{d_i}=0$ and $selection^j_{d_i}=0$ if operator $d_i$ was not selected in iteration $j$). The parameter $\lambda \in [0,1]$ indicates how influential is the performance of the operator during the execution of the solver: high values of $\lambda$ maintain rather steady weights, whereas low values of $\lambda$ allow for larger changes in the weights' values based on the performance of the operators. We consider $3$ different values for $\lambda$ in our experiments, $\lambda \in \{0.25, 0.50, 0.75\}$, denoted as ALNS$_{0.25}$,  ALNS$_{0.50}$, and ALNS$_{0.75}$ respectively. In Figure \ref{fig:alns_baseline} it is shown that ALNS with different $\lambda$ values does not provide better solutions than LNS, and that among the different versions increasing $\lambda$ provides slightly better results. 

\subsubsection{Changing the initial solution and the acceptance criterion:} 
\begin{figure}[h]
\centering
\includegraphics[width=0.5\linewidth]{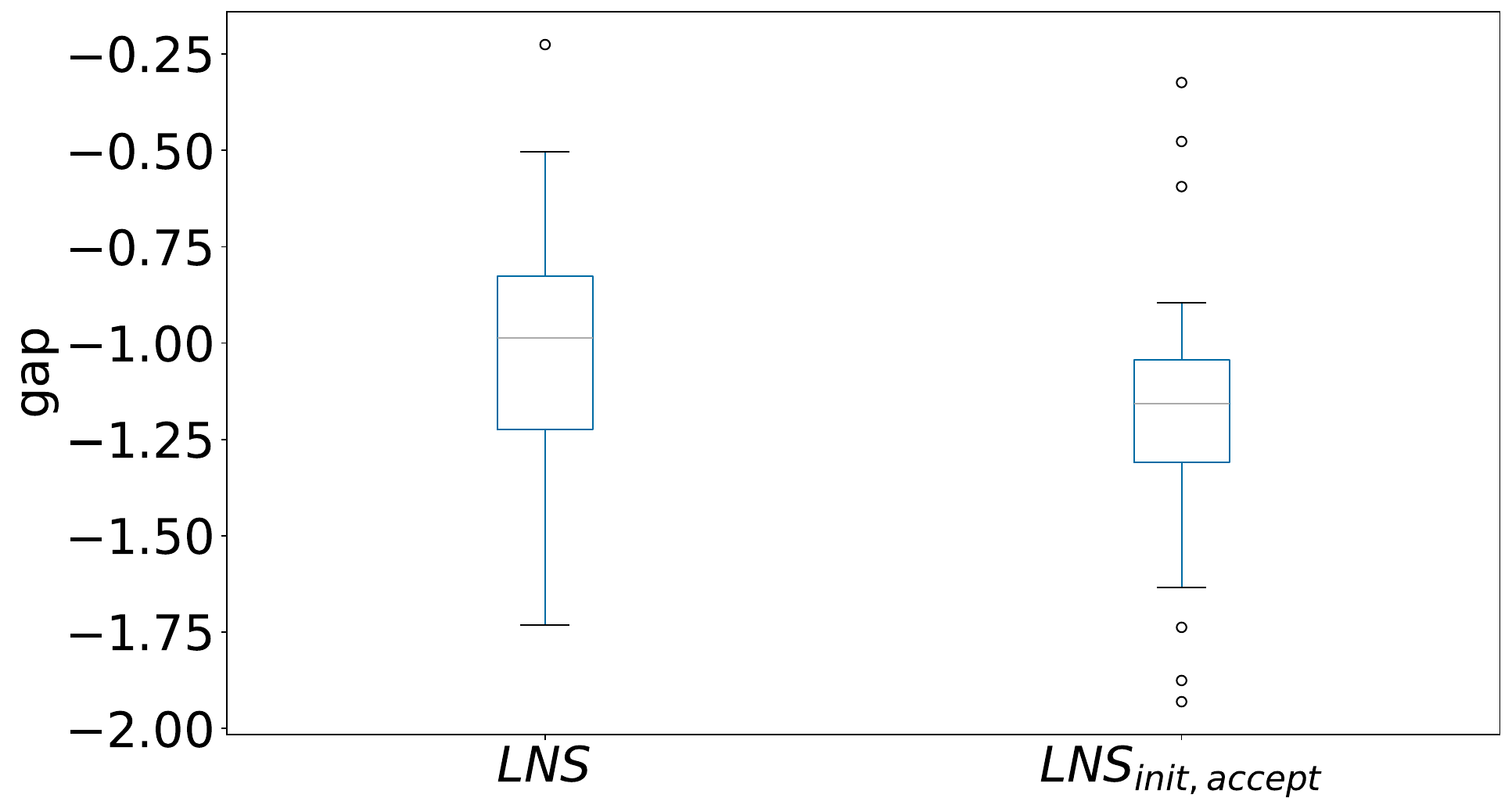}
\caption{GAP values for LNS and LNS$_{init,accept}$}
\label{fig:init_accept_baseline} 
\end{figure}

LNS$_{init,accept}$. Since changing the initial solution and the acceptance criterion independently has resulted in better solutions, we test another configuration where these changes are made simultaneously. As shown in Figure \ref{fig:init_accept_baseline}, this configuration yields better results than LNS.

\begin{figure}[h]
\centering
\includegraphics[width=0.5\linewidth, height=3cm]{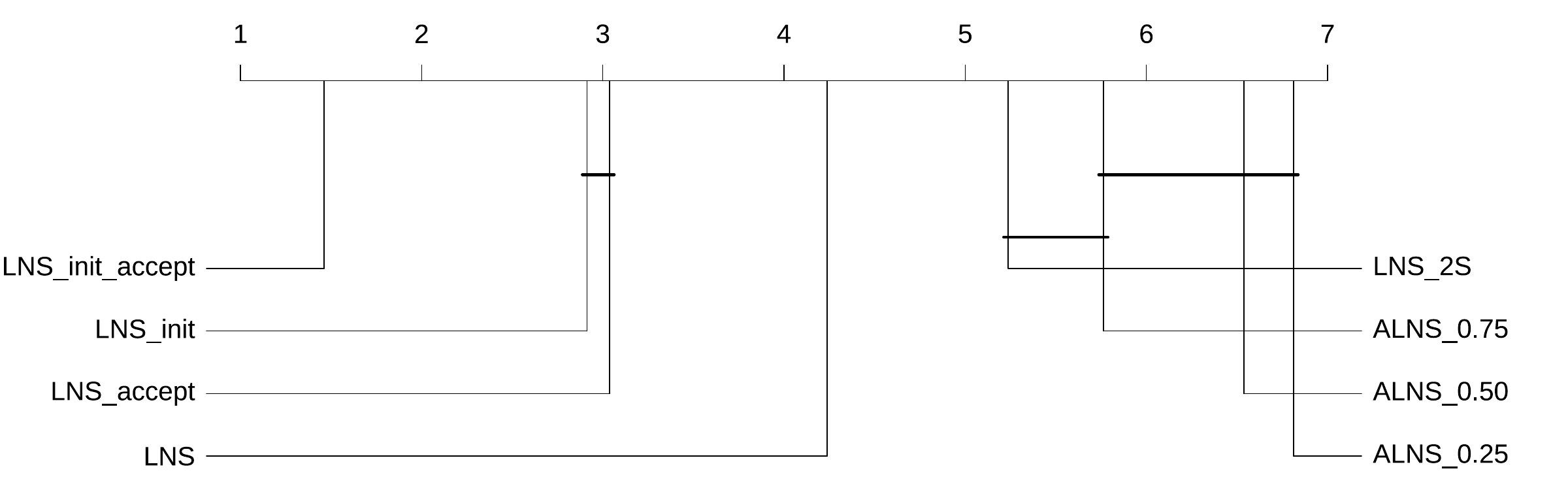}
\caption{Critical difference plot for different configurations}
\label{fig:configs_cd_plot}
\end{figure}

In order to provide a full comparison among the tested configurations, we also perform statistical tests using the R script \texttt{scmamp} \cite{calvo2016scmamp}. Firstly, the Friedman test shows that the evaluated configurations do not have the same performance, as the obtained p-value by this test is $2.2 \times 10^{-16}$. Furthermore, the results of the Nemenyi test are presented in Figure \ref{fig:configs_cd_plot}. These configurations are ranked based on their performance as follows: $LNS_{init,accept}$, $LNS_{init}$, $LNS_{accept}$, $LNS$, $LNS_{2S}$, $ALNS_{0.75}$, $ALNS_{0.50}$, and $ALNS_{0.25}$. As LNS$_{init,accept}$ is the first ranked configuration with significant difference, we use this algorithm for the comparison with state-of-the-art approaches.

\subsection{Results for Dataset \texttt{wlp}}
For the dataset \texttt{wlp} we compare the results of MR-MS-ILS, GRASP, PcEA, MG, SA, and our proposed LNS$_{init,accept}$. The experiments have been carried out for two different timeouts: $10\sqrt{m}$ and $m$ seconds.
\begin{figure}[h]
\centering
\begin{subfigure}[b]{0.45\textwidth}
\includegraphics[width=\textwidth]{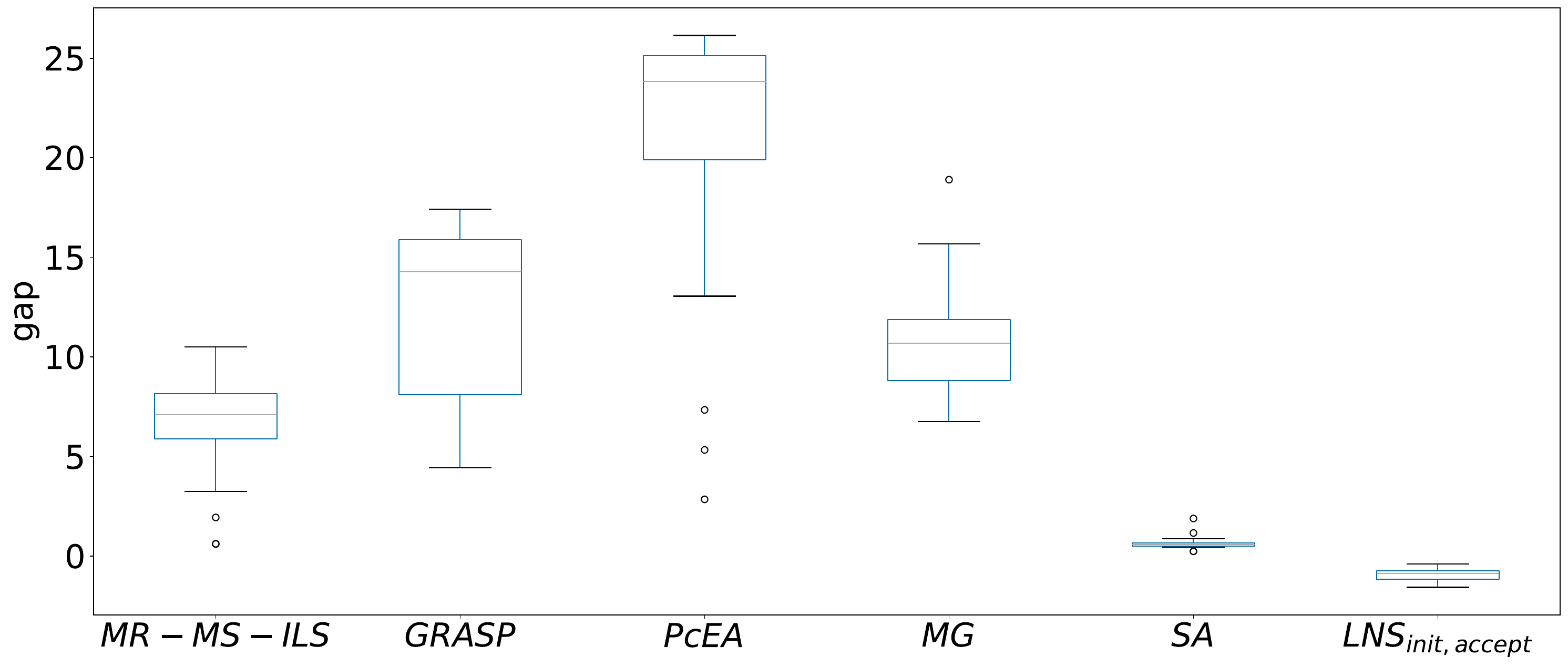}
\caption{GAP values}
\label{fig:gap_short_wlp}
\end{subfigure}
\hspace{5mm}
\begin{subfigure}[b]{0.45\textwidth}
\includegraphics[width=\textwidth]{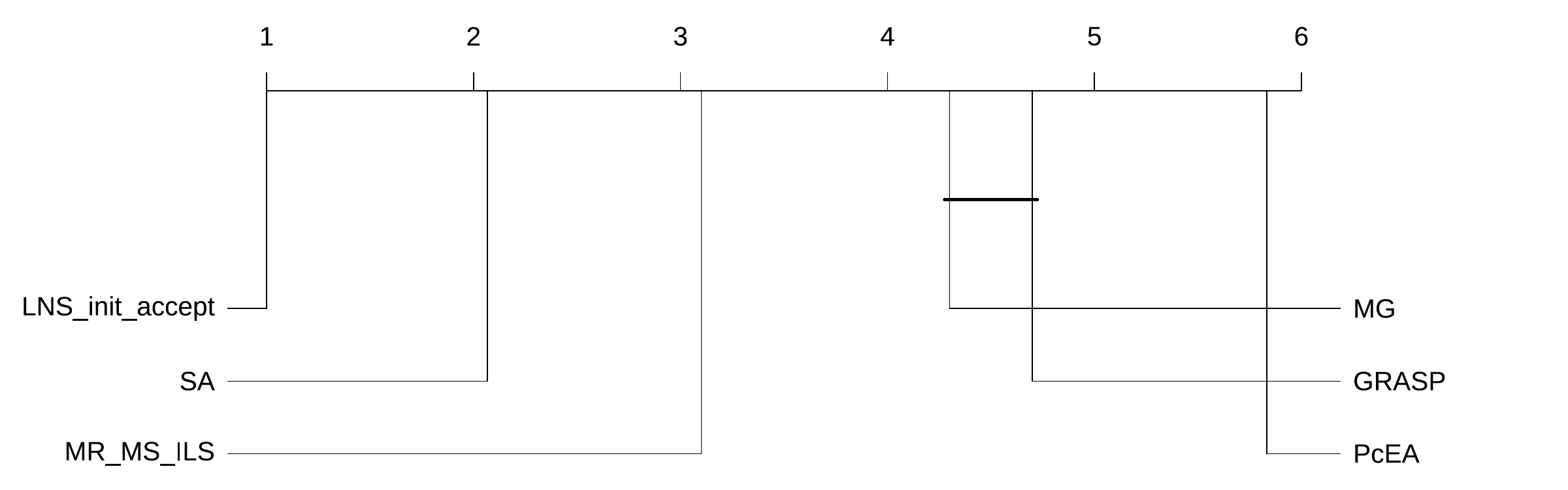}
\caption{Critical difference plot}
\label{fig:cd_short_wlp}
\end{subfigure}
\caption{Comparison with state of the art, dataset \texttt{wlp}, short timeout}
\end{figure}
From Figure \ref{fig:gap_short_wlp}, where the results for the shorter timeout are presented, the worst performing algorithm is PcEA. MR-MS-ILS, GRASP, and MG are also clearly outperformed by SA and LNS$_{init,accept}$. These two methods show very good and very consistent results, with LNS$_{init,accept}$ further outperforming SA on all instances. 
The critical difference plot in Figure \ref{fig:cd_short_wlp} confirms this picture, showing that LNS$_{init,accept}$ significantly improves results over SA, which is in turn significantly better than all other methods in comparison.

\begin{figure}[h]
\centering
\begin{subfigure}[b]{0.45\textwidth}
\includegraphics[width=\textwidth]{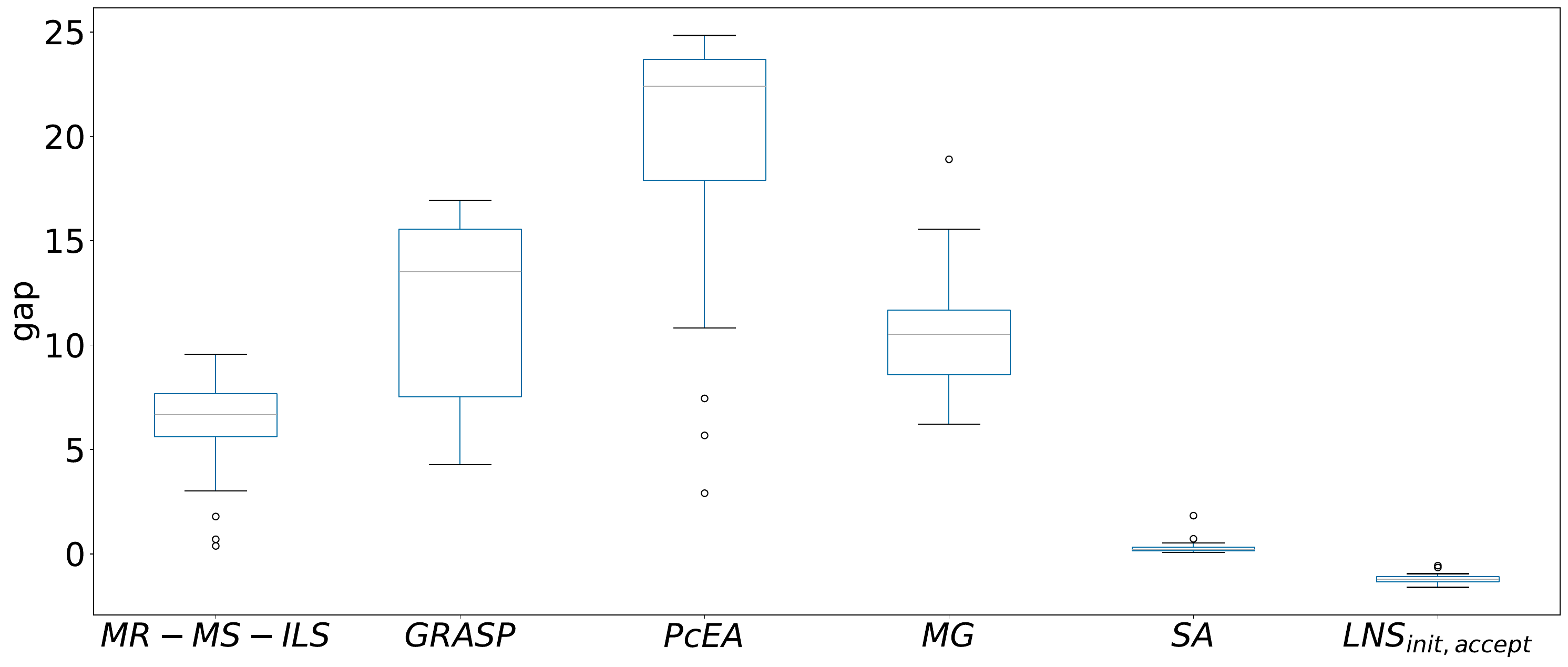}
\caption{GAP values}
\label{fig:gap_long_wlp}
\end{subfigure}
\hspace{5mm}
\begin{subfigure}[b]{0.45\textwidth}
\includegraphics[width=\textwidth]{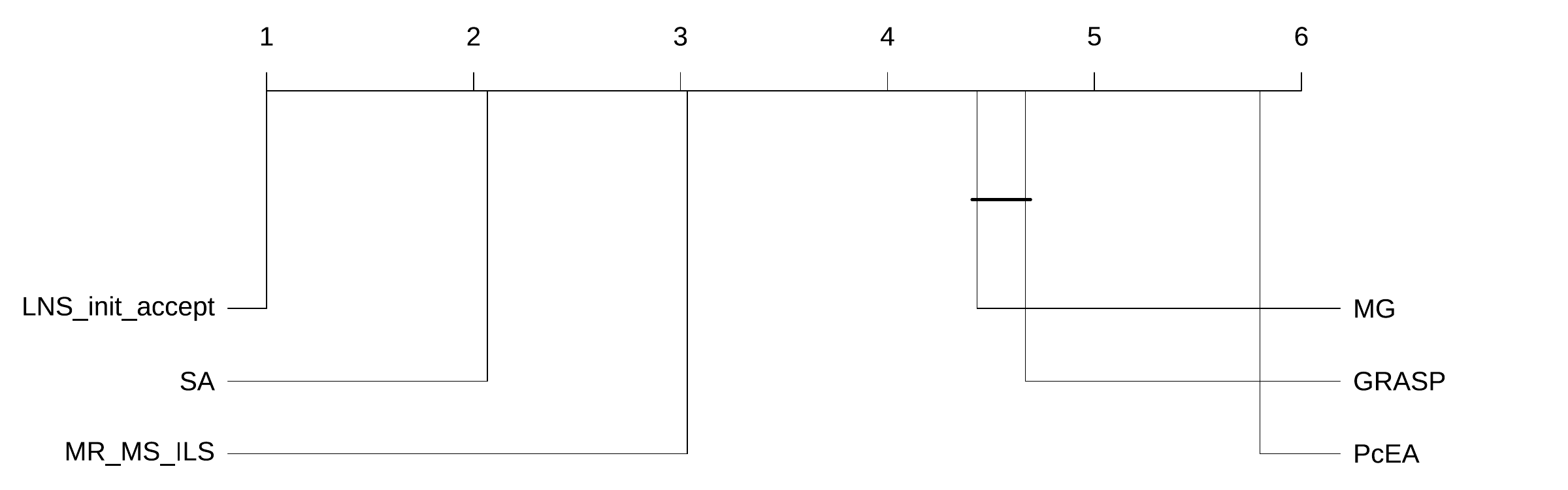}
\caption{Critical difference plot}
\label{fig:cd_long_wlp}
\end{subfigure}
\caption{Comparison with state of the art, dataset \texttt{wlp}, long timeout}
\end{figure}

Similar results are obtained using the longer timeout demonstrated in Figure \ref{fig:gap_long_wlp}. LNS$_{init,accept}$ consistently yields gap values below $0$ for all instances, which indicate that new best solutions have been found. Again, we perform statistical tests for all these methods based on the average reported results using  \texttt{scmamp}. The critical difference plot displayed in Figure \ref{fig:cd_long_wlp} shows that LNS$_{init,accept}$ significantly outperforms all other methods.

We present additional data for comparing SA and LNS$_{init,accept}$ in Table \ref{tab:wlp_tabular_results}, based on the long timeouts. We do not include the results from MR-MS-ILS, GRASP, PcEA, and MG as they are worse than SA (only MR-MS-ILS provides better results for $2$ instances compared to SA). In Table \ref{tab:wlp_tabular_results}, the first column shows the group of instances and the second column indicates the amount of entities involved in the respective group of instances. The number of entities is the sum of the number of facilities and the number of customers in an instance. The number of instances present in each group is given in the third column. Then, in the remaining columns we display the average results for each group of instances for the minimum and average costs obtained over $10$ runs for both SA and LNS$_{init,accept}$. For all the $4$ subgroups of instances, LNS$_{init,accept}$ gives better results for the minimum and average values. 

Actually, LNS$_{init,accept}$ finds new best solutions for all instances in dataset \texttt{wlp}. From the minimum values of LNS$_{init,accept}$ using the $m$ seconds timeout, across all instances, the minimum improvement in the gap value is $-0.68\%$ and the maximum improvement in the gap value is $-1.84\%$.

\begin{table*}[h]
\caption{Comparing SA and LNS$_{init,accept}$ for dataset \texttt{wlp}}
\label{tab:wlp_tabular_results}
\centering
\begin{tabular}{cccrrrr}
\toprule
& & & \multicolumn{2}{c}{SA} & \multicolumn{2}{c}{LNS$_{init,accept}$} \\
\cmidrule(lr){4-5} \cmidrule(lr){6-7}
group & entities & instances & \multicolumn{1}{c}{min} & \multicolumn{1}{c}{avg} & \multicolumn{1}{c}{min} & \multicolumn{1}{c}{avg}\\
\midrule
$1$ & $<1000$ & $8$ &  $72026.88$ & $72266.31$ & \textbf{70910.50} & \textbf{71056.31} \\ 
$2$ & $ [1000, 3000)$ & $9$ & $215053.44$ & $215532.24$ & \textbf{212022.67} & \textbf{212329.00}\\ 
$3$ & $ [3000, 6000)$ & $7$ & $412187.71$ & $412960.76$ & \textbf{407169.43} & \textbf{407631.50} \\ 
$4$ & $>=6000$ & $6$& $738703.00$ & $739522.45$ & \textbf{729244.00} & \textbf{729829.68}\\ 	
\bottomrule
\end{tabular}
\end{table*}

\subsection{Results for Dataset \texttt{cflp-ci}}

Dataset \texttt{cflp-ci}, composed of $50$ instances, was introduced by  \citet{ceschia2024multi}, therefore only SA was tested previously; and so we only compare SA and LNS$_{init,accept}$. 

\begin{figure}[h]
\centering
\begin{subfigure}{0.48\textwidth}
\includegraphics[width=\textwidth]{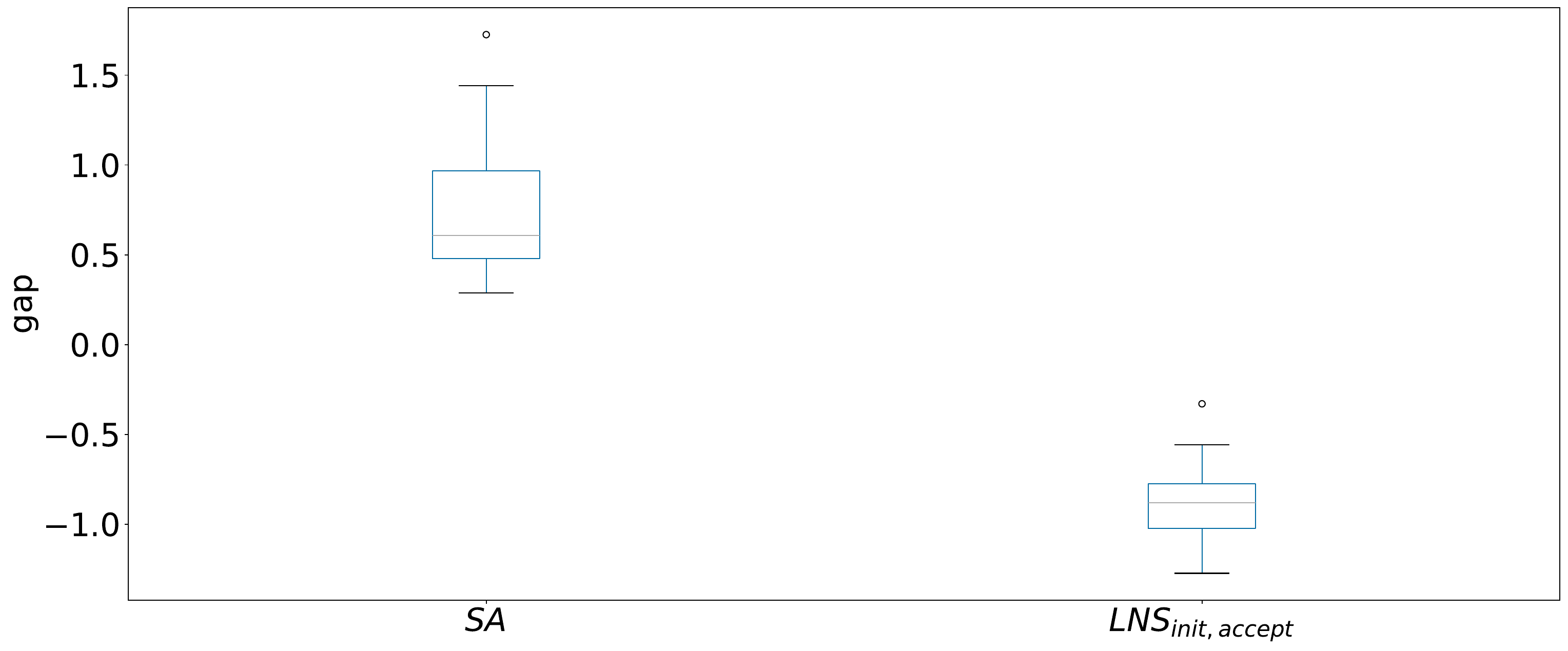}
\caption{GAP values, short timeout}
\label{fig:cflp_short_timeout}
\end{subfigure}
\hfill
\begin{subfigure}{0.48\textwidth}
\includegraphics[width=\textwidth]{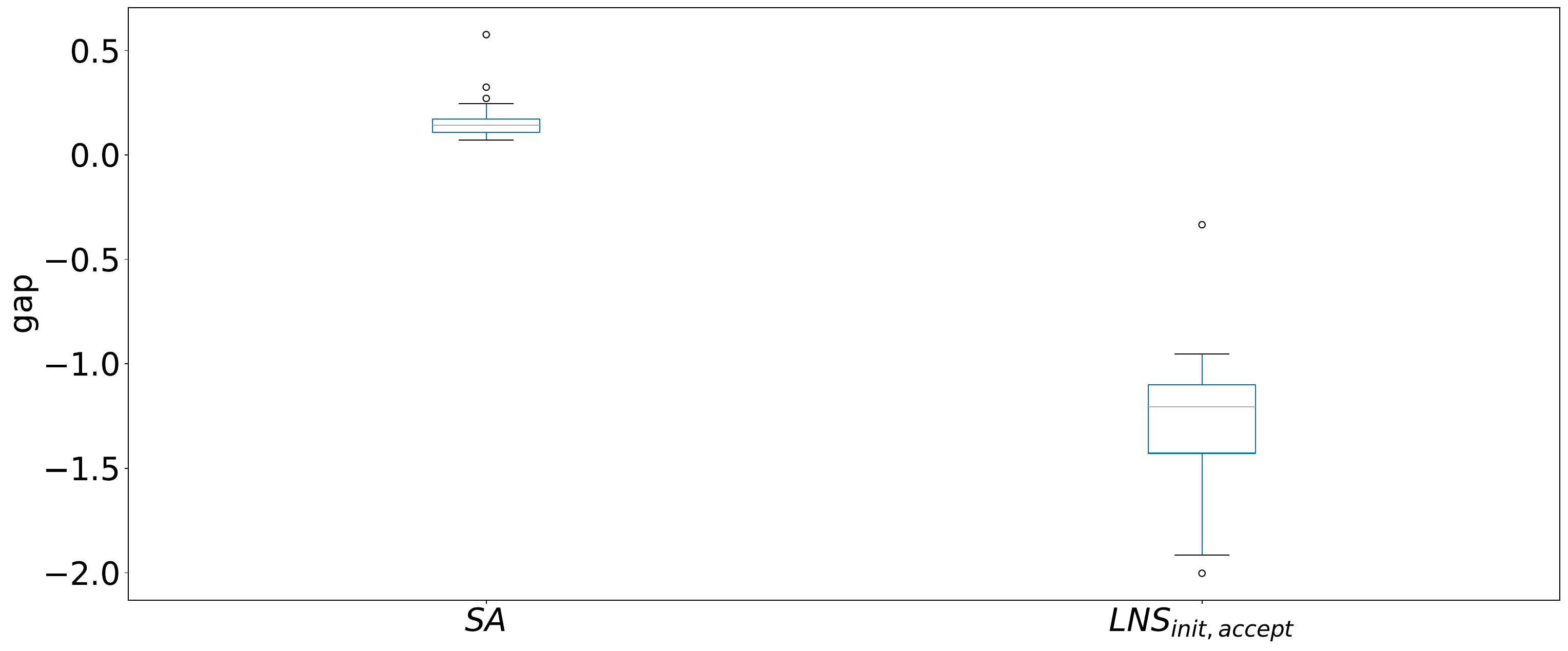}
\caption{GAP values, long timeout}
\label{fig:cflp_long_timeout}
\end{subfigure}
\caption{Comparison with state of the art, dataset \texttt{cflp-ci}}
\end{figure}

Again, in the shorter timeout as in Figure \ref{fig:cflp_short_timeout}, LNS$_{init,accept}$ is better than SA for all instances. We perform Wilcoxon signed-rank
test using the \texttt{scipy} module in python. With a significance level $\alpha=0.05$, a p-value $1.78 \times 10^{-15}$ is obtained. Such results confirm that LNS$_{init,accept}$ significantly outperforms SA.
Again, the improvement of LNS$_{init,accept}$ is even stronger in Figure \ref{fig:cflp_long_timeout}.

Similar to results of dataset \texttt{wlp}, we also compare SA and LNS$_{init,accept}$ based on the average results of the grouped instances based on the number of entities. In Table \ref{tab:cflp_tabular_results}, the first column shows the instance group, the second column shows the number of entities in the corresponding group and the third column gives the number of instances in each group. Note that in dataset \texttt{cflp-ci}, there are more large instances. Again, LNS$_{init,accept}$ has a superior performance compared to SA in terms of both minimum and average results. LNS$_{init,accept}$ provides new best solutions for all the instances of this second dataset as well. Based on the minimum values gained from LNS$_{init,accept}$ using the $m$ seconds timeout, across all instances, the minimum improvement in the gap value is $-0.46
\%$ and the maximum improvement in the gap value is $-2.09\%$. 
\begin{table*}[h]
\caption{Comparing SA and LNS$_{init,accept}$ for dataset \texttt{cflp-ci}}
\label{tab:cflp_tabular_results}
\centering
\begin{tabular}{cccrrrr}
\toprule 
& & & \multicolumn{2}{c}{SA} & \multicolumn{2}{c}{LNS$_{init,accept}$} \\
\cmidrule(lr){4-5} \cmidrule(lr){6-7}
group & entities & instances & \multicolumn{1}{c}{min} & \multicolumn{1}{c}{avg} & \multicolumn{1}{c}{min} & \multicolumn{1}{c}{avg}\\
\midrule
$1$ & $<3000$ & $10$ & $299350.90$ & $299924.00$ & \textbf{295640.20}& \textbf{295960.63}\\ 	
$2$ & $ [3000, 6000)$ & $11$ & $567097.09$ & $567913.34$ & \textbf{559708.18}& \textbf{560153.31}\\ 
$3$ & $ [6000, 8000)$ & $15$ & $650426.40$ & $651420.77$ & \textbf{640969.33} & \textbf{641535.73}\\ 		
$4$ & $>=8000$& $14$& $1119661.21$& $1120901.21$ & \textbf{1105434.36}& \textbf{1106188.05} \\ 	
\bottomrule
\end{tabular}
\end{table*}

Moreover, we compare the performance of SA and LNS$_{init,accept}$ for the instances of the second dataset, long timeout by again performing Wilcoxon signed-rank
test. With a significance level $\alpha=0.05$ the obtained p-value $1.78 \times 10^{-15}$ shows that LNS$_{init,accept}$ significantly outperforms SA.


\subsection{Further Analysis and Discussion}
\begin{figure}[h]
\centering
\includegraphics[width=0.5\textwidth]{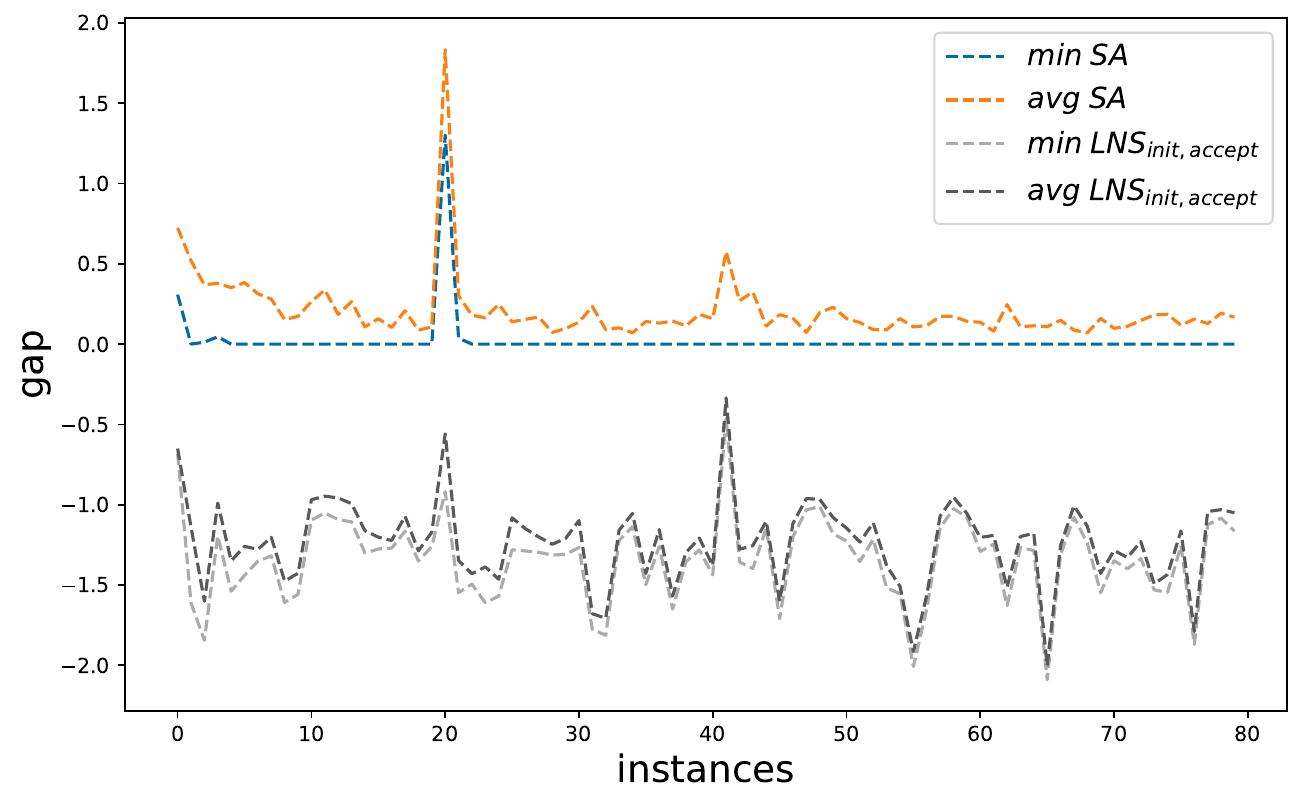}
\caption{GAP values for all instances}
\label{fig:gap_sa_lns}
\end{figure}
Further details are given in Figure \ref{fig:gap_sa_lns} to compare SA and LNS$_{init,accept}$ for all instances. The data of this graph represents gap values of the average and the minimum results from $10$ runs of SA and LNS$_{init,accept}$ each, where the BKS are given from the minimum values obtained by SA (only $2$ are from MR-MS-ILS). SA has slightly worse performance in small instances, and larger gaps between minimum and average in general. On the other hand, LNS$_{init,accept}$ seems to handle all of the instances well, which shows a very good performance. In comparison, it seems that the advantage of LNS$_{init,accept}$ over SA grows for larger runtimes. Detailed results for dataset \texttt{wlp} and for dataset \texttt{cflp-ci} can be found in the Appendix~\ref{sec:appendix}. The neighborhoods used in the SA perform small changes iteratively. In contrast, LNS$_{init,accept}$ explores mid-sized neighborhoods. This could help to further improve solutions while SA has already reached a local optimum. The tuning procedure regarding the sub-problem size was essential for the high performance of LNS$_{init,accept}$. 

\begin{figure}[h]
\includegraphics[width=0.5\textwidth]{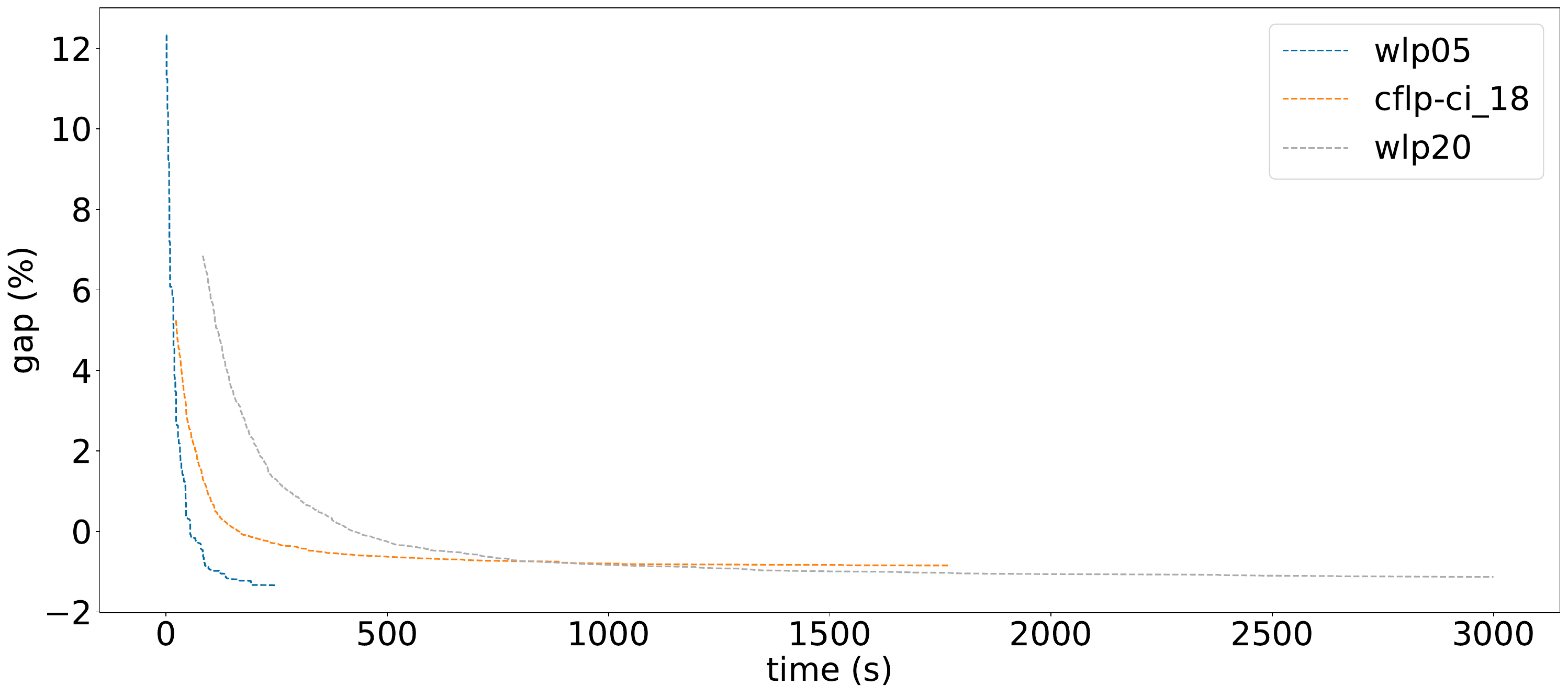}
\caption{GAP values over time for 3 different instances}
\label{fig:gap_values_over_time}
\end{figure}

In Figure \ref{fig:gap_values_over_time} we display the changes of the gap values over time for $3$  instances with different sizes: \texttt{wlp05}, \texttt{cflp-ci-18}, and \texttt{wlp20}. Each of these instances has a distinct timeout, which is based on the corresponding number of facilities. Note that in the Figure \ref{fig:gap_values_over_time} the threshold of a gap of 0 (the previous state-of-the-art solution) is reached before half the given runtime. As shown previously, LNS$_{init,accept}$ could reach better solutions already in the $10 \sqrt{m}$ seconds timeout.

Moreover, let us consider the data obtained from Gurobi in Table \ref{tab:exact_solvers} for the smaller instances. Comparing the minimum gap values of LNS$_{init,accept}$ with respect to the bounds given from Gurobi, the maximum gap value is $0.33\%$, whilst the minimum one is $0.00\%$ (LNS$_{init,accept}$ reaches the optimal solutions for two small instances). Recall that the time given to Gurobi was $7200$ seconds and LNS$_{init,accept}$ operates with $m$ seconds time budget. This shows how efficient LNS$_{init,accept}$ is using the given runtime compared to the application of exact methods.

\section{Conclusions}
\label{sec:conclusions}
In this paper, we present a Large Neighborhood Search for the multi-source capacitated facility location problem with customer incompatibilities. The proposed method includes novel destruction operators and a repair step based on an exact solver that optimizes selected sub-problems in each iteration. Our work provides a detailed investigation of algorithmic components, the configuration of our method, and analysis that includes statistical tests. The experimental results show that cheapest facilities \texttt{CF} and hybrid customers \texttt{HC} operators are especially useful for obtaining good solutions. Additionally, the repair operator has proven to be very effective, allowing the method to explore medium to large neighborhoods in a short time. Furthermore, integrating another initial solution heuristic and changing the acceptance criteria has enhanced the performance of the baseline algorithm. Overall, our evaluation demonstrates that our method currently represents the state of the art for this problem and can be used to solve large instances in a reasonable time. 

A possible direction for future work would be to perform an Instance Space Analysis \cite{smith2023instance} in order to understand which are the features that make a given instance hard for each specific method. In particular, using this analysis, we may identify new instances where LNS still outperforms other techniques, as well as instances that are more challenging for LNS. To conclude, we plan to try to extend our method to other versions of the facility location problem, and compare with the state-of-the-art results for the available benchmarks for these variants.


\bibliographystyle{ACM-Reference-Format}
\bibliography{refs}

\newpage
\appendix
\section{Appendices}
\label{sec:appendix}

\begin{table*}[h]
\caption{Detailed performance results of SA and LNS (variant: LNS$_{init,accept}$) on the \texttt{wlp} dataset under both timeout settings.}
\label{tab:detailed_results_wlp}
\centering
{\fontsize{9}{12}\selectfont
\begin{tabular}{lrrrrrrrr}
\toprule
& \multicolumn{4}{c}{$10\sqrt{m}$ seconds timeout} & \multicolumn{4}{c}{$m$ seconds timeout} \\
\cmidrule(lr){2-5} 
\cmidrule(lr){6-9} 
 & \multicolumn{2}{c}{SA} & \multicolumn{2}{c}{LNS} & \multicolumn{2}{c}{SA} & \multicolumn{2}{c}{LNS} \\
 \cmidrule(lr){2-3} \cmidrule(lr){4-5} \cmidrule(lr){6-7} \cmidrule(lr){8-9}
instance & \multicolumn{1}{c}{min} & \multicolumn{1}{c}{avg} & \multicolumn{1}{c}{min} & \multicolumn{1}{c}{avg} & \multicolumn{1}{c}{min} & \multicolumn{1}{c}{avg} & \multicolumn{1}{c}{min} & \multicolumn{1}{c}{avg} \\
\midrule
wlp01 & 29025.0 & 29249.7 & \bfseries 28716.0 & \bfseries 28731.0 & 29002.0 & 29122.3 & \bfseries 28716.0 & \bfseries 28725.0 \\
wlp02 & 54061.0 & 54207.5 & \bfseries 52967.0 & \bfseries 53180.5 & 53838.0 & 54117.5 & \bfseries 52970.0 & \bfseries 53224.4 \\
wlp03 & 65562.0 & 65986.2 & \bfseries 64370.0 & \bfseries 64532.3 & 65570.0 & 65804.1 & \bfseries 64353.0 & \bfseries 64512.3 \\
wlp04 & 85894.0 & 86391.7 & \bfseries 84717.0 & \bfseries 85010.3 & 85933.0 & 86219.4 & \bfseries 84868.0 & \bfseries 85043.5 \\
wlp05 & 106079.0 & 106336.0 & \bfseries 104260.0 & \bfseries 104524.6 & 105814.0 & 106185.8 & \bfseries 104186.0 & \bfseries 104386.2 \\
wlp06 & 113976.0 & 114255.3 & \bfseries 111867.0 & \bfseries 112328.6 & 113499.0 & 113934.0 & \bfseries 111865.0 & \bfseries 112069.4 \\
wlp07 & 165647.0 & 166063.1 & \bfseries 162814.0 & \bfseries 163189.0 & 165105.0 & 165623.4 & \bfseries 162874.0 & \bfseries 162995.0 \\
wlp08 & 191822.0 & 192275.7 & \bfseries 188883.0 & \bfseries 189121.0 & 191002.0 & 191537.5 & \bfseries 188483.0 & \bfseries 188704.6 \\
wlp09 & 222979.0 & 223537.7 & \bfseries 219348.0 & \bfseries 219737.6 & 222498.0 & 222838.3 & \bfseries 218919.0 & \bfseries 219214.3 \\
wlp10 & 249762.0 & 250453.9 & \bfseries 245854.0 & \bfseries 246281.6 & 249199.0 & 249629.1 & \bfseries 245317.0 & \bfseries 245643.1 \\
wlp11 & 294315.0 & 294877.0 & \bfseries 290544.0 & \bfseries 290947.1 & 293349.0 & 294125.8 & \bfseries 290135.0 & \bfseries 290506.6 \\
wlp12 & 302834.0 & 303594.2 & \bfseries 299271.0 & \bfseries 299617.0 & 301602.0 & 302619.9 & \bfseries 298430.0 & \bfseries 298746.9 \\
wlp13 & 320652.0 & 321310.4 & \bfseries 316879.0 & \bfseries 317539.7 & 319647.0 & 320238.1 & \bfseries 316155.0 & \bfseries 316585.1 \\
wlp14 & 402477.0 & 403560.0 & \bfseries 397520.0 & \bfseries 397952.1 & 400871.0 & 401929.2 & \bfseries 396436.0 & \bfseries 396889.7 \\
wlp15 & 466848.0 & 467414.7 & \bfseries 460270.0 & \bfseries 460785.2 & 464710.0 & 465208.8 & \bfseries 458666.0 & \bfseries 459311.0 \\
wlp16 & 541385.0 & 542326.1 & \bfseries 533837.0 & \bfseries 534884.1 & 539320.0 & 540173.0 & \bfseries 532439.0 & \bfseries 532838.3 \\
wlp17 & 573244.0 & 574713.0 & \bfseries 565930.0 & \bfseries 566605.8 & 571361.0 & 571954.6 & \bfseries 564102.0 & \bfseries 564373.4 \\
wlp18 & 638976.0 & 640471.3 & \bfseries 631258.0 & \bfseries 631856.1 & 636129.0 & 637451.0 & \bfseries 628717.0 & \bfseries 629312.1 \\
wlp19 & 757538.0 & 758421.8 & \bfseries 747526.0 & \bfseries 748279.9 & 754102.0 & 754760.4 & \bfseries 743917.0 & \bfseries 744396.2 \\
wlp20 & 994310.0 & 995029.6 & \bfseries 980351.0 & \bfseries 980994.8 & 986397.0 & 987448.3 & \bfseries 973960.0 & \bfseries 974845.9 \\
wlp21 & 38872.0 & 39147.1 & \bfseries 38176.0 & \bfseries 38270.1 & 38920.0 & 39124.3 & \bfseries 38067.0 & \bfseries 38205.3 \\
wlp22 & 75860.0 & 76043.7 & \bfseries 74651.0 & \bfseries 74823.6 & 75888.0 & 76088.7 & \bfseries 74686.0 & \bfseries 74837.6 \\
wlp23 & 121275.0 & 121540.2 & \bfseries 119398.0 & \bfseries 119673.4 & 121250.0 & 121468.4 & \bfseries 119438.0 & \bfseries 119516.2 \\
wlp24 & 172252.0 & 172672.6 & \bfseries 169553.0 & \bfseries 169816.7 & 171887.0 & 172168.9 & \bfseries 169118.0 & \bfseries 169502.6 \\
wlp25 & 234307.0 & 235019.8 & \bfseries 230745.0 & \bfseries 231296.9 & 234000.0 & 234580.6 & \bfseries 230328.0 & \bfseries 230573.7 \\
wlp26 & 295774.0 & 296392.6 & \bfseries 292049.0 & \bfseries 292333.5 & 294942.0 & 295352.6 & \bfseries 291165.0 & \bfseries 291751.7 \\
wlp27 & 397231.0 & 397998.0 & \bfseries 392040.0 & \bfseries 392527.4 & 395901.0 & 396510.9 & \bfseries 390802.0 & \bfseries 391349.7 \\
wlp28 & 465282.0 & 466056.0 & \bfseries 458867.0 & \bfseries 459473.8 & 463263.0 & 464045.4 & \bfseries 457258.0 & \bfseries 457699.8 \\
wlp29 & 603826.0 & 605266.3 & \bfseries 596421.0 & \bfseries 596953.1 & 602169.0 & 602607.9 & \bfseries 594258.0 & \bfseries 594663.2 \\
wlp30 & 888124.0 & 888870.3 & \bfseries 875615.0 & \bfseries 876313.2 & 882060.0 & 882912.5 & \bfseries 870510.0 & \bfseries 871387.3 \\
\bottomrule
\end{tabular}
}
\end{table*}
\begin{table*}
\caption{Detailed performance results of SA and LNS (variant: LNS$_{init,accept}$) on the \texttt{cflp-ci} dataset under both timeout settings.}
\label{tab:detailed_results_cflp_part1}
\centering
\resizebox{\textwidth}{!}{
{\fontsize{9}{12}\selectfont
\begin{tabular}{lrrrrrrrr}
\toprule
& \multicolumn{4}{c}{$10\sqrt{m}$ seconds timeout} & \multicolumn{4}{c}{$m$ seconds timeout} \\
\cmidrule(lr){2-5} 
\cmidrule(lr){6-9} 
 & \multicolumn{2}{c}{SA} & \multicolumn{2}{c}{LNS} & \multicolumn{2}{c}{SA} & \multicolumn{2}{c}{LNS} \\
 \cmidrule(lr){2-3} \cmidrule(lr){4-5} \cmidrule(lr){6-7} \cmidrule(lr){8-9}
instance & \multicolumn{1}{c}{min} & \multicolumn{1}{c}{avg} & \multicolumn{1}{c}{min} & \multicolumn{1}{c}{avg} & \multicolumn{1}{c}{min} & \multicolumn{1}{c}{avg} & \multicolumn{1}{c}{min} & \multicolumn{1}{c}{avg} \\
\midrule
cflp-ci\_00 & 424519.0 & 425264.1 & \bfseries 418823.0 & \bfseries 419410.5 & 423418.0 & 423997.0 & \bfseries 418054.0 & \bfseries 418765.8 \\
cflp-ci\_01 & 573983.0 & 575397.2 & \bfseries 562163.0 & \bfseries 562799.3 & 568919.0 & 570251.9 & \bfseries 558816.0 & \bfseries 559368.4 \\
cflp-ci\_02 & 448326.0 & 449302.0 & \bfseries 438197.0 & \bfseries 439013.9 & 442916.0 & 443314.1 & \bfseries 434885.0 & \bfseries 435353.2 \\
cflp-ci\_03 & 1508473.0 & 1509804.8 & \bfseries 1484309.0 & \bfseries 1485992.0 & 1495309.0 & 1496823.6 & \bfseries 1477031.0 & \bfseries 1477974.5 \\
cflp-ci\_04 & 1220618.0 & 1222512.9 & \bfseries 1205899.0 & \bfseries 1206907.3 & 1216587.0 & 1217446.0 & \bfseries 1202747.0 & \bfseries 1203756.5 \\
cflp-ci\_05 & 368456.0 & 368914.3 & \bfseries 360889.0 & \bfseries 361370.9 & 364036.0 & 364548.5 & \bfseries 358581.0 & \bfseries 358836.8 \\
cflp-ci\_06 & 155149.0 & 155376.4 & \bfseries 153236.0 & \bfseries 153459.6 & 154811.0 & 155013.3 & \bfseries 152860.0 & \bfseries 153021.8 \\
cflp-ci\_07 & 452561.0 & 453707.1 & \bfseries 443153.0 & \bfseries 443722.0 & 447859.0 & 448501.9 & \bfseries 440471.0 & \bfseries 440836.3 \\
cflp-ci\_08 & 702000.0 & 703657.0 & \bfseries 691660.0 & \bfseries 692597.4 & 699317.0 & 700112.5 & \bfseries 689844.0 & \bfseries 690242.9 \\
cflp-ci\_09 & 334128.0 & 334818.7 & \bfseries 328938.0 & \bfseries 329204.0 & 331089.0 & 331702.6 & \bfseries 326845.0 & \bfseries 327089.8 \\
cflp-ci\_10 & 612854.0 & 614279.3 & \bfseries 603666.0 & \bfseries 604039.8 & 610333.0 & 611297.3 & \bfseries 601560.0 & \bfseries 601912.5 \\
cflp-ci\_11 & 30894.0 & 30975.9 & \bfseries 30739.0 & \bfseries 30780.3 & 30882.0 & 31059.8 & \bfseries 30739.0 & \bfseries 30778.8 \\
cflp-ci\_12 & 461171.0 & 462177.8 & \bfseries 454589.0 & \bfseries 455108.5 & 460229.0 & 461474.0 & \bfseries 453984.0 & \bfseries 454351.4 \\
cflp-ci\_13 & 388264.0 & 388879.6 & \bfseries 382291.0 & \bfseries 382656.3 & 386948.0 & 388202.4 & \bfseries 381543.0 & \bfseries 382092.8 \\
cflp-ci\_14 & 1031054.0 & 1032521.9 & \bfseries 1017459.0 & \bfseries 1018150.8 & 1027028.0 & 1028186.6 & \bfseries 1015264.0 & \bfseries 1015706.4 \\
cflp-ci\_15 & 565607.0 & 566459.6 & \bfseries 553905.0 & \bfseries 554673.3 & 561099.0 & 562128.0 & \bfseries 551504.0 & \bfseries 552110.0 \\
cflp-ci\_16 & 891160.0 & 892011.8 & \bfseries 879714.0 & \bfseries 880076.7 & 887877.0 & 889296.2 & \bfseries 877228.0 & \bfseries 877976.0 \\
cflp-ci\_17 & 1231972.0 & 1233547.8 & \bfseries 1216612.0 & \bfseries 1217444.7 & 1225297.0 & 1226184.4 & \bfseries 1212645.0 & \bfseries 1213514.8 \\
cflp-ci\_18 & 903653.0 & 905310.9 & \bfseries 893100.0 & \bfseries 893962.8 & 900438.0 & 902206.0 & \bfseries 891335.0 & \bfseries 891733.0 \\
cflp-ci\_19 & 152349.0 & 152740.9 & \bfseries 150418.0 & \bfseries 150633.9 & 151907.0 & 152255.5 & \bfseries 150113.0 & \bfseries 150265.8 \\
cflp-ci\_20 & 1280618.0 & 1283255.5 & \bfseries 1263127.0 & \bfseries 1264067.7 & 1273308.0 & 1275338.5 & \bfseries 1257699.0 & \bfseries 1258724.3 \\
cflp-ci\_21 & 460143.0 & 461168.5 & \bfseries 453867.0 & \bfseries 454769.2 & 459288.0 & 459903.4 & \bfseries 453069.0 & \bfseries 453624.8 \\
cflp-ci\_22 & 313629.0 & 314280.3 & \bfseries 309909.0 & \bfseries 310312.3 & 313381.0 & 313663.0 & \bfseries 309574.0 & \bfseries 309895.2 \\
cflp-ci\_23 & 414336.0 & 415852.6 & \bfseries 408925.0 & \bfseries 409350.0 & 413761.0 & 414130.3 & \bfseries 407480.0 & \bfseries 408052.1 \\
cflp-ci\_24 & 473660.0 & 474247.5 & \bfseries 463703.0 & \bfseries 464289.5 & 468618.0 & 469362.6 & \bfseries 461333.0 & \bfseries 461549.4 \\
cflp-ci\_25 & 803729.0 & 805523.5 & \bfseries 786013.0 & \bfseries 786602.4 & 796370.0 & 797221.8 & \bfseries 780385.0 & \bfseries 781119.9 \\
cflp-ci\_26 & 698430.0 & 700168.0 & \bfseries 687043.0 & \bfseries 687553.1 & 695323.0 & 696126.2 & \bfseries 683861.0 & \bfseries 684456.3 \\
cflp-ci\_27 & 1433262.0 & 1434453.2 & \bfseries 1411200.0 & \bfseries 1412939.1 & 1420856.0 & 1423304.2 & \bfseries 1404596.0 & \bfseries 1405652.9 \\
cflp-ci\_28 & 469428.0 & 470734.6 & \bfseries 464260.0 & \bfseries 465099.3 & 469035.0 & 469843.7 & \bfseries 464232.0 & \bfseries 464564.5 \\
cflp-ci\_29 & 1381347.0 & 1383402.3 & \bfseries 1362356.0 & \bfseries 1363975.1 & 1372452.0 & 1374401.2 & \bfseries 1357651.0 & \bfseries 1357976.4 \\
cflp-ci\_30 & 580934.0 & 581995.5 & \bfseries 572891.0 & \bfseries 573628.0 & 579263.0 & 580050.7 & \bfseries 571785.0 & \bfseries 572289.8 \\
cflp-ci\_31 & 1216734.0 & 1218218.8 & \bfseries 1196202.0 & \bfseries 1197591.3 & 1206339.0 & 1207307.3 & \bfseries 1191308.0 & \bfseries 1191970.6 \\
cflp-ci\_32 & 492232.0 & 493294.5 & \bfseries 481729.0 & \bfseries 482570.7 & 487236.0 & 488433.6 & \bfseries 479301.0 & \bfseries 479854.8 \\
cflp-ci\_33 & 1082543.0 & 1083618.0 & \bfseries 1065711.0 & \bfseries 1066578.2 & 1075666.0 & 1076818.6 & \bfseries 1062045.0 & \bfseries 1062779.3 \\
cflp-ci\_34 & 1156148.0 & 1158156.4 & \bfseries 1141153.0 & \bfseries 1142128.5 & 1151861.0 & 1153176.1 & \bfseries 1137058.0 & \bfseries 1138285.4 \\
cflp-ci\_35 & 549152.0 & 550971.0 & \bfseries 534835.0 & \bfseries 535629.3 & 541626.0 & 542213.1 & \bfseries 530308.0 & \bfseries 530779.2 \\
cflp-ci\_36 & 556058.0 & 557097.0 & \bfseries 548433.0 & \bfseries 548888.6 & 554259.0 & 555083.6 & \bfseries 546993.0 & \bfseries 547305.3 \\
cflp-ci\_37 & 1290114.0 & 1291258.3 & \bfseries 1273849.0 & \bfseries 1274930.3 & 1284102.0 & 1285221.9 & \bfseries 1270259.0 & \bfseries 1271169.0 \\
cflp-ci\_38 & 1058405.0 & 1059980.9 & \bfseries 1044581.0 & \bfseries 1046002.8 & 1054933.0 & 1055679.1 & \bfseries 1041865.0 & \bfseries 1042975.9 \\
cflp-ci\_39 & 192846.0 & 193320.9 & \bfseries 190002.0 & \bfseries 190416.8 & 192693.0 & 193001.8 & \bfseries 189710.0 & \bfseries 189941.7 \\
\bottomrule
\end{tabular}
}
}
\end{table*}
\begin{table*}
\caption{Detailed performance results of SA and LNS (variant: LNS$_{init,accept}$) on the \texttt{cflp-ci} dataset under both timeout settings.}
\label{tab:detailed_results_cflp_part2}
\centering
\resizebox{\textwidth}{!}{
{\fontsize{9}{12}\selectfont
\begin{tabular}{lrrrrrrrr}
\toprule
& \multicolumn{4}{c}{$10\sqrt{m}$ seconds timeout} & \multicolumn{4}{c}{$m$ seconds timeout} \\
\cmidrule(lr){2-5} 
\cmidrule(lr){6-9} 
 & \multicolumn{2}{c}{SA} & \multicolumn{2}{c}{LNS} & \multicolumn{2}{c}{SA} & \multicolumn{2}{c}{LNS} \\
 \cmidrule(lr){2-3} \cmidrule(lr){4-5} \cmidrule(lr){6-7} \cmidrule(lr){8-9}
instance & \multicolumn{1}{c}{min} & \multicolumn{1}{c}{avg} & \multicolumn{1}{c}{min} & \multicolumn{1}{c}{avg} & \multicolumn{1}{c}{min} & \multicolumn{1}{c}{avg} & \multicolumn{1}{c}{min} & \multicolumn{1}{c}{avg} \\
\midrule
cflp-ci\_40 & 638996.0 & 639737.2 & \bfseries 629472.0 & \bfseries 630335.0 & 636674.0 & 637300.3 & \bfseries 628087.0 & \bfseries 628495.9 \\
cflp-ci\_41 & 1309439.0 & 1311166.0 & \bfseries 1287787.0 & \bfseries 1288993.5 & 1297401.0 & 1298835.6 & \bfseries 1279258.0 & \bfseries 1280155.7 \\
cflp-ci\_42 & 422885.0 & 424160.1 & \bfseries 417425.0 & \bfseries 417777.1 & 422267.0 & 422889.1 & \bfseries 416627.0 & \bfseries 417074.7 \\
cflp-ci\_43 & 508871.0 & 509989.0 & \bfseries 499698.0 & \bfseries 500148.0 & 505548.0 & 506465.7 & \bfseries 497811.0 & \bfseries 498000.6 \\
cflp-ci\_44 & 766352.0 & 767928.0 & \bfseries 751923.0 & \bfseries 752678.8 & 759920.0 & 761332.5 & \bfseries 748174.0 & \bfseries 749018.3 \\
cflp-ci\_45 & 724761.0 & 725291.0 & \bfseries 714139.0 & \bfseries 714861.3 & 721010.0 & 721857.7 & \bfseries 711966.0 & \bfseries 712621.5 \\
cflp-ci\_46 & 637155.0 & 638433.8 & \bfseries 625334.0 & \bfseries 625658.2 & 633722.0 & 634707.6 & \bfseries 621870.0 & \bfseries 622387.5 \\
cflp-ci\_47 & 410913.0 & 411767.5 & \bfseries 406255.0 & \bfseries 406792.3 & 410205.0 & 410729.5 & \bfseries 405593.0 & \bfseries 405928.5 \\
cflp-ci\_48 & 778044.0 & 779471.1 & \bfseries 769013.0 & \bfseries 769655.1 & 775605.0 & 777095.8 & \bfseries 767200.0 & \bfseries 767610.7 \\
cflp-ci\_49 & 305726.0 & 306143.6 & \bfseries 300998.0 & \bfseries 301598.1 & 304209.0 & 304719.1 & \bfseries 300666.0 & \bfseries 301013.6 \\
\bottomrule
\end{tabular}
}
}
\end{table*}

\end{document}